\pdfoutput=1
\documentclass[format=acmsmall, review=false, screen=true]{acmart}
\AtBeginDocument{%
  \providecommand\BibTeX{{%
    \normalfont B\kern-0.5em{\scshape i\kern-0.25em b}\kern-0.8em\TeX}}}

\setcopyright{acmcopyright}
\copyrightyear{2018}
\acmYear{2018}
\acmDOI{XXXXXXX.XXXXXXX}

\acmConference[Conference acronym 'XX]{Make sure to enter the correct
  conference title from your rights confirmation emai}{June 03--05,
  2018}{Woodstock, NY}
\acmPrice{15.00}
\acmISBN{978-1-4503-XXXX-X/18/06}

\newtheorem{problem}{Problem}
\usepackage{booktabs} 
\usepackage{subfigure} 
\usepackage{url}
\usepackage{tabu}
\usepackage{array}
\usepackage{xspace,amsmath,multirow,color,array,colortbl}
\usepackage{natbib}
\usepackage{lineno} \linenumbers

\usepackage{makecell} 

\renewcommand\footnotetextcopyrightpermission[1]{}
\usepackage{bbding}
\usepackage[misc]{ifsym}
\usepackage{threeparttable}
\usepackage{ragged2e}
\usepackage[printonlyused]{acronym}
\renewcommand{\raggedright}{\leftskip=0pt \rightskip=0pt plus 0cm}

\acrodef{DNN}{\emph{Deep Neural Network}} 
\acrodef{LSTMs}{\emph{Long Short-Term Memory Recurrent Networks}}
\acrodef{CNN}{\emph{Convolution Neural Network}}
\acrodef{RNN}{\emph{Recurrent Neural Network}}
\acrodef{GCRN}{\emph{Graph Convolutional Recurrent Network}}
\acrodef{GCNs}{\emph{Graph Convolutional Networks}} 
\acrodef{GNN}{\emph{Graph Neural Network}} 
\acrodef{GAT}{\emph{Graph ATtention network}} 
\acrodef{PLM}{\emph{Pre-trained Language Model}} 
\acrodef{MLP}{\emph{Multi-Layer Perceptron}} 
\acrodef{BERT}{\emph{Bidirectional Encoder Representations from Transformers}}
\acrodef{NLP}{\emph{Natural Language Processing}} 
\acrodef{KG}{\emph{Knowledge Graph}} 

\begin{document}
\switchlinenumbers	
\nolinenumbers


\title[Multi-turn Response Selection]{Multi-turn Response Selection with Commonsense-enhanced Language Models}

\author{Yuandong Wang}
\affiliation{
	\institution{Tsinghua University}
	    \city{Haidian District}
	    \state{Beijing}
	\country{China}
}
\email{wangyd2021@tsinghua.edu.cn}

\author{Xuhui Ren}
\affiliation{
	\institution{The University of Queensland}
	    \city{Brisbane}
	    \state{QLD}
	\country{Australia}
}
\email{xuhui.ren@uq.net.au}

\author{Tong Chen}
\affiliation{
	\institution{The University of Queensland}
	    \city{Brisbane}
	    \state{QLD}
	\country{Australia}
}
\email{tong.chen@uq.edu.au}

\author{Yuxiao Dong}
\affiliation{
	\institution{Tsinghua University}
	    \city{Haidian District}
	    \state{Beijing}
	\country{China}
}
\email{yuxiaod@tsinghua.edu.cn}

\author{Nguyen Quoc Viet Hung}
\affiliation{%
	\institution{Griffith University}
	    \city{Brisbane}
	    \state{QLD}
	\country{Australia}
}
\email{henry.nguyen@griffith.edu.au}

\author{Jie Tang*}
\affiliation{
	\institution{Tsinghua University}
	    \city{Haidian District}
	    \state{Beijing}
	\country{China}
}
\email{jietang@tsinghua.edu.cn}

\thanks{* Corresponding author.}
\renewcommand{\shortauthors}{Wang and Tang, et al.}

\begin{abstract}
As a branch of advanced artificial intelligence, dialogue systems are prospering. Multi-turn response selection is a general research problem in dialogue systems. With the assistance of background information and pre-trained language models, the performance of state-of-the-art methods on this problem gains impressive improvement. However, existing studies neglect the importance of external commonsense knowledge. Hence, we design a \underline{\textbf{Si}}amese \underline{\textbf{n}}etwork where a pre-trained \underline{\textbf{L}}anguage model merges with a \underline{\textbf{G}}raph neural network (\textbf{SinLG}). SinLG takes advantage of Pre-trained Language Models (PLMs) to catch the word correlations in the context and response candidates and utilizes a Graph Neural Network (GNN) to reason helpful common sense from an external knowledge graph. The GNN aims to assist the PLM in fine-tuning, and arousing its related memories to attain better performance. Specifically, we first extract related concepts as nodes from an external knowledge graph to construct a subgraph with the context response pair as a super node for each sample. Next, we learn two representations for the context response pair via both the PLM and GNN. A similarity loss between the two representations is utilized to transfer the commonsense knowledge from the GNN to the PLM. Then only the PLM is used to infer online so that efficiency can be guaranteed. Finally, we conduct extensive experiments on two variants of the PERSONA-CHAT dataset, which proves that our solution can not only improve the PLM’s performance but also achieve an efficient inference.
\end{abstract}

%



\maketitle
\section{Introduction}

\begin{figure*}[htbp]
	\setlength{\abovecaptionskip}{0.2cm} 
	\setlength{\belowcaptionskip}{-0.2cm}
	\centering
	\includegraphics[width=0.8\textwidth]{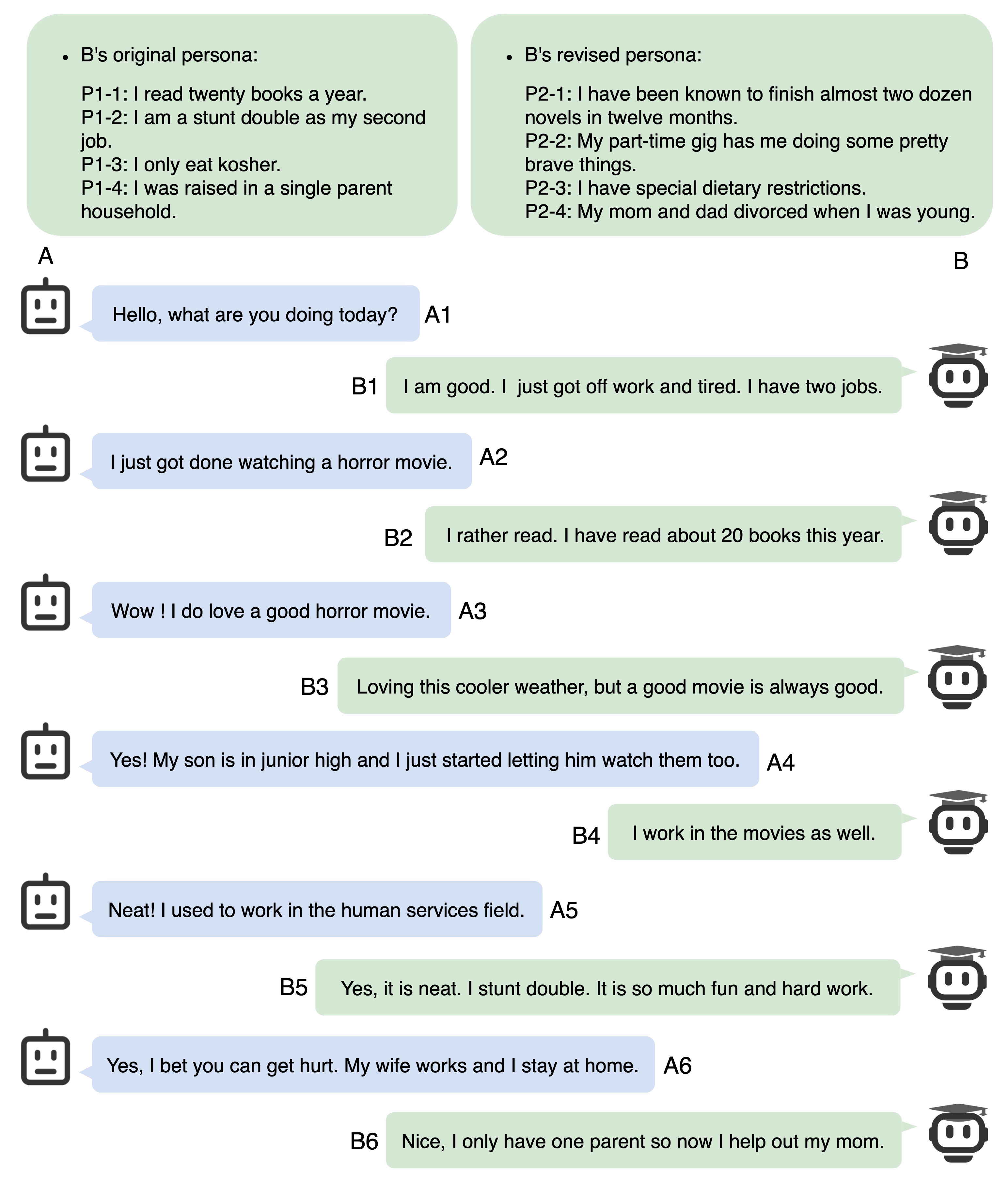}
	\caption{An example for multi-turn conversations with persona information. For convenience, we provide each utterance in the dialogue with an identifier, such as P1-1 for the first item of B's original persona, A1 for the first utterance of A. The revised persona is rephrased from the original one in an implicit way, which is more challenging for the dialogue agent to comprehend and respond.}
	\label{fig:dialogue-ex}
\end{figure*}
One of the most significant tasks of advanced artificial intelligence has been building a dialogue system possessing the cognitive ability which can communicate with humans coherently. Recent dialogue systems can be categorized into 
generation-based and retrieval-based according to the methodologies applied. 
Retrieval-based dialogue systems are usually given a dialog corpus and the user's current utterance, and then a small response candidate set would be returned via a retrieval algorithm while generation-based methods often produce a diverse response candidate set first. Then both kinds of dialogue systems come to the situation where a response candidate set would be obtained before responding to users. Unfortunately, not all response candidates are rational under the current dialogue context. Hence, it is very necessary to provide matching scores for these response candidates and choose the best ones. Obviously, obtaining accurate matching scores for response candidates lies at the core of most dialogue systems. 

In this paper, we investigate the specific problem named \textbf{M}ulti-turn \textbf{R}esponse \textbf{S}election (\textbf{MRS}) which evaluates and filters response candidates for multi-turn dialogues. 
Traditional methods focus on generating more comprehensive representations for the context and response separately and obtaining the final matching score via a matching layer. Specifically, hand-crafted features are utilized at first to provide clues for the selection of the best response, e.g., sentence length, the number of common words for sentence-level representation~\cite{wang2017deep} and topic information~\cite{wu2018response}. With the prosperity of deep neural networks, researchers employ \ac{CNN}, \ac{RNN}, and self-attention mechanisms coupled with \ac{MLP} and pooling layers to learn effective representations~\cite{yan2018response}. 
Under the multi-turn scenario, it is vital to effectively catch the relationships between different sentences in the context, as well as those between contexts and response candidates to yield a comprehensive embedding of them. A multi-turn dialogue system is required to be capable of following topic changes from time to time and returning fluent responses to users. For instance, \textbf{Fig.\ref{fig:dialogue-ex}} shows a multi-turn dialogue between A and B from PERSONA-CHAT~\cite{zhang2018personalizing}. In the first round, they are talking about B's current state and job; then the topic transfers to their hobbies and families in the following rounds gradually. Therefore, we need to consider a multi-turn context, so as to determine which response candidate is more suitable for the current context. 
However, the above methods may catch the relationships between different sentences in the context but fail to conduct sufficient interactions among the context and response candidates, which makes it hard to learn the relationships among the context and response candidates comprehensively. 

As both the context and response candidates carry complex semantics, diverse new frameworks~\cite{zhou2018multi,tao2019one,yuan2019multi} are presented to mine the interactions between them so that the matching accuracy can be improved. The recent \ac{PLM}s~\cite{devlin2019bert, liu2019roberta} exploit the interactions between all context sentences and response candidates via the attention mechanism in a more comprehensive way. Variants of \ac{PLM}-based models~\cite{henderson2019training, whang2020effective, xu2021learning} for MRS are proposed recently. Particularly,~\cite{henderson2019training} learns representations for the context and response separately via \ac{BERT} and aggregates them to compute the final matching score. Further,~\citet{xu2021learning} design several self-supervised tasks and jointly pre-train the \ac{PLM}-based model with them for MRS in a multi-task manner. As the application scenarios of dialogue systems are diversifying, MRS involves more complicated information, e.g., background or persona restrictions. 
Hence, there has been an increasing need for moving beyond learning only the dialogue features, towards capturing relationships between such complicated context and response candidates.

Simultaneously, various methodologies~\cite{zhao2019document, gu2020filtering,liu2020k, zhang2021adapting, zhu2021content} are proposed to employ background information for MRS, such as profiles of users and entity infoboxes from Wikipedia. 
For instance,~\citet{gu2020filtering} propose FIRE which filters context and knowledge at first, and their embeddings are generated severally by bidirectional and global attention mechanisms. 
\citet{liu2020k} incorporate knowledge triplets into the utterances and feed them into BERT while they use soft position and visible matrix to restrain the effect of noise from the knowledge.
Unfortunately, these methods overlook a significant problem: how to select a rational response candidate consistent with common sense? Human beings talk to each other with common sense besides the given background knowledge, which is absent in the above methods. There is a lack of trustworthy reasoning from the background to response candidates. Referring to the example in \textbf{Fig.\ref{fig:dialogue-ex}}, B's persona is given in four sentences as a knowledge supplement. With P1-1 and P1-4, we can recognize B2 and B6 are the correct responses consistent with B's persona. It is relatively easy for both humans and a state-of-art model to associate the information about twenty books read in a year in both the persona and conversation, so as ``single parent’’ and ``one parent’’. However, the persona might not necessarily be as well-structured, where a typical example is shown in Fig.1 (marked as B's revised persona). In this case, ``two dozen’’ in P2-1 and ``about 20" in B2, ``mom and dad divorced’’ in P2-4, and ``one parent’’ in B6 immediately make it challenging for a language model to precisely match the context without guidance from commonsense knowledge. Therefore, trustworthy reasoning ability usually needs to be equipped with external common sense which can help understand the relevance between background information and response candidates so as to enhance the performance of models on the MRS task. In addition, complex MRS tasks like persona-based chat tend to be more demanding on sophisticated and labor-intensive annotations. Consequently, with the substantially limited training data, a method that can take advantage of commonsense reasoning would be highly desirable for optimizing the conversation responses selected.

In view of the challenges and deficiencies of existing studies, we design a framework called \underline{\textbf{Si}}amese \underline{\textbf{n}}etwork combining pre-trained \underline{\textbf{L}}anguage model with \underline{\textbf{G}}raph neural network (i.e., \textbf{SinLG}). As is well-known, \ac{PLM}s can accumulate lots of language principles and knowledge according to the pre-training process via its powerful memorization ability from enormous parameters. \ac{PLM}s possess the strong capability of language representation and understanding, but its performance gain compared with previous studies might be restrained because of improper applications, especially when the target task needs specific background knowledge. Therefore, we propose to enhance the performance of \ac{PLM}s by incorporating common sense from an external knowledge graph with a \ac{GNN} so that the related knowledge memories of \ac{PLM}s can be aroused. The \ac{GNN} is responsible for augmenting useful commonsense information from the extra knowledge graph and assists the \ac{PLM} in fine-tuning. With the supplement of commonsense knowledge, the performance of \ac{PLM}s can be improved even on more difficult understanding tasks. 
Instead of appending the representation learned from a \ac{KG} after the \ac{PLM}'s representation directly, we propose the \ac{KG} guided training for efficient inference where a similarity-based and self-supervised objective function is optimized to help \ac{PLM}s achieve better performance. In this way, we can not only transfer the commonsense knowledge from the \ac{GNN} to \ac{PLM}s but also augment the supervision signal, which enables our framework to generalize better amid the limited training data.
By this means, the \ac{PLM} can better exploit its strong capability of language representation and understanding to model the dialogue text where the multi-head attention mechanism can fully catch the relationships between each utterance in the context, as well as response candidates to yield a comprehensive understanding for the response selection. Meanwhile, the heavy computations in the \ac{GNN} part can be omitted during the inference, such as entity linking, concept ranking, and so on, which is so time-consuming that would lead to high latency and poor user experience.

The main contributions of this work are summarized as follows:
\begin{itemize}
	
	\item Considering the challenges and insufficiencies of state-of-the-art approaches to the MRS problem, we propose to supplement common sense from the external knowledge graph to \ac{PLM}s so that their performance can be enhanced by specific background information. \vspace{1mm}
	
	\item In order to optimize the inference efficiency, we propose the KG-guided training which employs a similarity-based and self-supervised objective function to arouse the related knowledge memories of \ac{PLM}s to obtain a better performance. \vspace{1mm}
	
	\item We conduct extensive experiments on different variants of a public dataset, which testifies the superiority of our proposed solution from several different perspectives.\vspace{1mm}

\end{itemize}

The paper is organized as follows. Section \ref{sec:pre} introduces some basic definitions of MRS problem. The details of our solution on MRS are presented in Section \ref{sec:solution}. In Section \ref{sec:evaluation}, we provide extensive experimental results to analyze the effectiveness of our model from various perspectives. Section \ref{sec:related} reviews the development of existing related works and Section \ref{sec:conclusion} concludes the paper and proposes the prospects and limitations of this work. 

\section{Problem Statement}\label{sec:pre}

In this section, we define some basic notations and mathematically formulate our problem. 
\subsection{Definitions}\label{sec:definitions}

\begin{definition}\label{def:w}
	\textit{\textbf{Token $w$.}} In order to process the natural language into codes that computers can understand, the common operation usually divides them into word-by-word pieces, called tokens which can map into ids as inputs of models. Here we denote a token as $w$ in this paper.
\end{definition}

\begin{definition}\label{def:u}
\textit{\textbf{Utterance $u$.}} An utterance $u$ is a chronological sequence of tokens and usually expresses a complete thought, which is denoted as $u$ $=$ $\{w_1,$ $w_2,$ $...,$ $w_{|u|}\}$ where $|u|$ is the number of tokens in this utterance.

\end{definition}

\begin{definition}\label{def:P}
	\textit{\textbf{Persona $P$.}} Persona $P$ is a set of utterances that describe the profile of one side in a given dialogue. It provides background information for the dialogue. In particular, we define this kind of utterance as $p$ $=$ $\{w_1,$ $w_2,$ $...,$ $w_{|p|}\}$, and persona $P$ $=$ $\{p_1,$ $p_2,$ $...,$ $p_{|P|}\}$.
	
\end{definition}

\begin{definition}\label{def:C}
	\textit{\textbf{Context $C$.}} A dialogue's context $C$ is a chronological sequence of utterances, which is denoted as $C$ $=$ $\{u_1,$ $u_2,$ $...,$ $u_{|C|}\}$ where $|C|$ is the number of utterances in this context. 
	
\end{definition}

\begin{definition}\label{def:R}
	\textit{\textbf{Response Candidate Set $R$.}} According to the context, a dialogue system can obtain a series of utterances as response candidates. Specially, we represent a response candidate as $r$ which also can be denoted as $r$ $=$ $\{w_1,$ $w_2,$ $...,$ $w_{|r|}\}$.  Similar to the above, the response candidate set can be represented as $R$ $=$ $\{r_1,$ $r_2,$ $...,$ $r_{|R|}\}$.
	
\end{definition}

\begin{definition}\label{def:KG}
	\textit{\textbf{Knowledge Graph $\mathcal{G}$.}} We define the external knowledge graph that we extract common sense from as a multi-relational graph $\mathcal{G}$ $=$ $(\mathcal{V}$, $\mathcal{E}$, $\mathcal{R})$ where $\mathcal{V}$ is the set of entity/concept nodes (which are usually composed of a meaningful word or phrase. A token in dialogues can correspond to an entity in $\mathcal{G}$ during the knowledge extraction.), $\mathcal{E}$ $\subseteq$ $\mathcal{V}$ $\times$ $\mathcal{R}$ $\times$ $\mathcal{V}$ is the set of edges between nodes, and $\mathcal{R}$ is the set of relation types in $\mathcal{G}$.
\end{definition}

\subsection{Multi-turn Response Selection (\textbf{MRS})}
\begin{problem}\label{def:problem}
	\textbf{Multi-turn Response Selection (\textbf{MRS})}. Given a dialogue dataset $\mathcal{D}$ $=$ $\{D_i:$ $(P_i,$ $C_i,$ $R_i,$ $Y_i)\}_{i=1}^ {|\mathcal{D}|}$, where $Y_i$ $=$ $\{y_1,$ $y_2,$ $...,$ $y_{|Y_i|}$\} represents the rational scores of  response candidates in $R_i$ (note ${|Y_i|}$ $=$ ${|R_i|}$), and an external knowledge graph $\mathcal{G}$ $=$ $(\mathcal{V}, \mathcal{E}, \mathcal{R})$ with common sense, our goal is to learn a matching model $f(P_i,$ $C_i,$ $\mathcal{G},$ $R_i)$ $\to$ $Y_i$ that measures the relevance between $C_i$ and the candidates in $R_i$. In the following part of this paper, we refer to a sample $(P_i,$ $C_i,$ $R_i,$ $Y_i)\in \mathcal{D}$ as a \textbf{mrsSample}.
\end{problem}

\section{SOLUTION}\label{sec:solution}
\begin{figure}[tb!]
	\centering
	\setlength{\abovecaptionskip}{0.15cm}
	\subfigure[Training]{
		\label{fig:model-training}
		\includegraphics[width=0.40\textwidth]{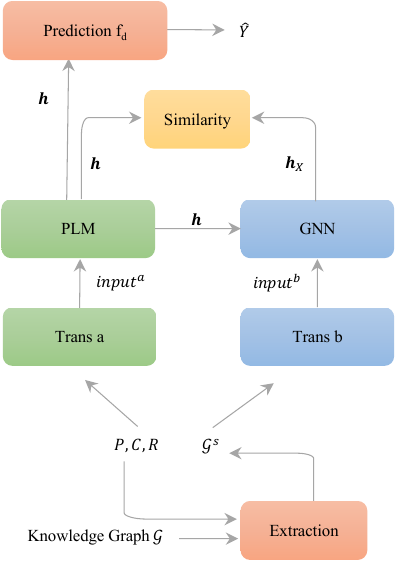}}
	\hspace{12ex}
	\subfigure[Inference]{
		\label{fig:model-inference}
		\includegraphics[width=0.15\textwidth]{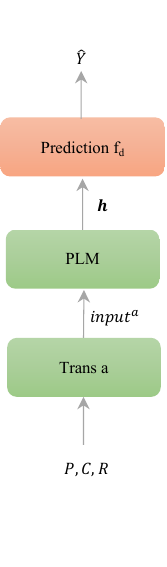}}
	
	\caption{An overview of our solution. Note this figure is consistent with the content from Section 3.1 to Section 3.4, i.e., \textbf{Eq.(\ref{eq:score})} to \textbf{Eq.(\ref{eq:final-loss})}. A dialogue text includes $P$, $C$, and $R$ which  denote persona, context, and response candidate set, respectively, as defined in Section \ref{sec:definitions}. $\mathcal{G}^s$ is the subgraph extracted from the knowledge graph $\mathcal{G}$. Trans a and b represent two different transformations of the input data and are given details in Section \ref{sec:2trans}. Prediction $f_d$ is a neural network layer to calculate the final results from the embedding vectors, e.g., a feed-forward layer.}
	\label{fig:model} 
\end{figure}

In order to incorporate common sense into the dialogue system, we design a framework which is a siamese network combining a pre-trained language model (PLM) with a graph neural network (GNN). An overview is shown in \textbf{Fig.\ref{fig:model}}.
The framework is inspired by both Siamese network architecture~\cite{chen2021exploring}, and the dual process theory from the cognitive process of humans~\cite{evans1984heuristic, evans2003two, evans2008dual, sloman1996empirical}. In the Siamese network architecture, a \ac{PLM} is employed to capture the relationships among different utterances of the context, as well as the relationships between the context and response candidates via its strong capability of language representation and understanding. Meanwhile, a \ac{GNN} is in place to gather useful common sense from the external knowledge graph and assists the PLM in the fine-tuning process. 
To endow the PLM with the common sense extracted by the GNN when selecting the next response, we additionally incorporate a term in the training loss that encourages the similarity between the representations generated by these two components. 
Through this framework, we can leave out the commonly adopted process~\cite{yasunaga2021qa, feng2020scalable} that extracts and ranks related nodes from the external knowledge graph during the inference, which is time-consuming and would lead to high latency and poor user experience. 

The whole technical process of SinLG is composed of the following steps. First, we extract relevant concepts from the knowledge graph $\mathcal{G}$ via linking and scoring. Specifically, entity linking is conducted between tokens in the conversation context and concepts in $\mathcal{G}$, and then a PLM is employed to score the abstracted concepts via computing the similarity between them and the corresponding context~\cite{yasunaga2021qa}. Combing the dialogue text with extracted relevant concepts as the model input, we conduct two different transformations on them to respectively generate the inputs of the \ac{PLM} and \ac{GNN}. Each of them learns a unique representation for the context-response pair. A similarity score between them is computed as one part of the final loss. Finally, the representation vector from the \ac{PLM} is fed into the prediction layer to obtain matching scores of context-response pairs.
The general form of \ac{PLM}s that we used in our framework is defined as a composition of two functions:
\begin{equation}
	\begin{split}
		\textbf{h} &= f_e(\textbf{u}), \\
		\hat{y} &= f_d(\textbf{h} ),
	\end{split}
\end{equation}
where encoder $f_e$ separates the utterance $\textbf{u}$ as tokens, represents it with a sequence of token ids, and learns a contextualized representation $\textbf{h}$ for $\textbf{u}$; then $f_d$ is a layer to fit for a desired task taking $\textbf{h}$ as its input. 

\subsection{Knowledge Extraction}
Dialogue systems, especially those open-domain ones, are expected to converse with humans in an engaging way, so common sense is very necessary to be integrated into the model effectively~\cite{young2018augmenting}. When coming across a specific context, dialogue systems are usually incapable of reasoning according to corresponding background information unless we empower them with specific means. This section aims to extract useful commonsense concepts from an external knowledge graph and rank them based on their relevance to the dialogue context. 

First of all, we need to select an external knowledge graph $\mathcal{G}$ which would include useful common sense as much as possible. As the \textbf{Fig.\ref{fig:model-training}} shows, we retrieve a subgraph $\mathcal{G}_{i,k}^s$ from $\mathcal{G}$ for each response candidate and its context, i.e., $(P_i,$ $C_i,$ $r_{i,k},$ $y_{i,k})$, following prior works~\cite{feng2020scalable, yasunaga2021qa}. The extraction process includes four steps. First, given $(P_i,$ $C_i,$ $r_{i,k},$ $y_{i,k})$, the concept set $\mathcal{V}_{i,k}^{PCR}$ is obtained by entity linking between tokens in the dialogue text and concept nodes in $\mathcal{G}$. Second, the few-hop neighbors of elements in $\mathcal{V}_{i,k}^{PCR}$ are extracted to form $\mathcal{V}_{i,k}^{e}$, which is a set of extra nodes containing high-order external knowledge from the knowledge graph. The complete node set of the subgraph is the union of the two, $\mathcal{V}_{i,k}^{s}$ $=$ $\mathcal{V}_{i,k}^{PCR}$ $\cup$ $\mathcal{V}_{i,k}^{e}$. Third, we calculate a score for each concept node in $\mathcal{V}_{i,k}^{s}$ to measure the relevance between the dialogue text and corresponding concept through the \ac{PLM}: 
\begin{equation}
	\label{eq:score}
	\begin{split}
		\textbf{h}_{i,k} &= f_e (P_i \oplus C_i \oplus r_{i,k} ),  \\
		score_j &= f_m(\textbf{h}_{i,k}, f_e (v_j)),  v_j \in \mathcal{V}_{i,k}^{s}, \\
		f_m(\boldsymbol{\mu}, \boldsymbol{\sigma}) &= \boldsymbol{\mu} \cdot \boldsymbol{\sigma}, 
	\end{split}
\end{equation}
where $\oplus$ and $\cdot$ represent concatenation and dot product, respectively, and $f_m$ is the dot-product similarity function. Since a longer average text length of multi-turn dialogues leads to more related concepts from $\mathcal{G}$, calculating such pairwise scores becomes computationally impractical for a large set of concepts. As the equations show, we employ a bi-encoder that first encodes the context and concepts into embeddings, and then computes their similarity scores through their embedding vectors. Compared with the cross-encoder, a bi-encoder is more convenient and efficient under our multi-turn scenario where there are more candidate concepts to score because of the longer context and it allows to pre-compute the embeddings of candidate concepts first ~\cite{humeau2019poly, li2021virt}. In practice, we pre-compute the embeddings of all concepts in $\mathcal{G}$ and conduct a dot product between the embedding of dialogue text and its corresponding concept embeddings to obtain their final scores. Both embeddings of the concepts and dialogue text are computed by the \ac{PLM}. Then we rank these concepts according to their similarity scores with the context and choose the top ones as useful inputs of the GNN. 
The embeddings of chosen top concepts will be updated via GNN’s message passing in \textbf{Eq.(\ref{eq:gnn})}.
Finally, the edges $\mathcal{E}_{i,k}^{s}$ and relation types $\mathcal{R}_{i,k}^{s}$ are abstracted based on $\mathcal{V}_{i,k}^{s}$, that is, $\mathcal{G}_{i,k}^{s}$ $=$ $(\mathcal{V}_{i,k}^{s}$, $\mathcal{E}_{i,k}^{s}$, $\mathcal{R}_{i,k}^{s})$.

\subsection{SinLG}\label{sec:sinlg}
To effectively inject common sense into the PLM for response selection, we utilize \ac{GNN} to learn the structural information from $\mathcal{G}$ in SinLG. Meanwhile, \ac{GNN} transfers the commonsense knowledge to the PLM by maximizing the similarity between the different embeddings from the \ac{PLM} and \ac{GNN} where the design of Siamese network structure aims to realize the augmentation effect for the same input. In this way, the performance of the \ac{PLM} can be enhanced. 

\subsubsection{\textbf{Two Transformations}}\label{sec:2trans}
Above all, besides the dialogue text $P_i,$ $C_i,$ $r_{i,k}$, the input of our model also includes its subgraph $\mathcal{G}_{i,k}^{s}$. The two components of SinLG manage different available information that they are good at. The PLM models the dialogue text so we propose \textbf{Trans a} to preprocess $P_i,$ $C_i,$ $r_{i,k}$, while the GNN models common sense from the \ac{KG} and \textbf{Trans b} is to deal with the graph-structured knowledge. 
In the left part of \textbf{Fig.\ref{fig:model-training}}, we conduct \textbf{Trans a} which means concatenating $P_i$, $C_i$, $r_{i,k}$ one by one:
\begin{equation}
	\label{eq:transa}
	\begin{split}
     \textit{input}^a_{i,k} &= P_i \oplus C_i \oplus r_{i,k} .
	\end{split}
\end{equation}
In the right part of \textbf{Fig.\ref{fig:model-training}}, a virtual and super node $X_{i,k}$ representing the context is created and added into $\mathcal{G}_{i,k}^{s}$. 
The relationships between $X_{i,k}$ and all other nodes are measured by their scores calculated via \textbf{Eq.(\ref{eq:score})}. Then we assume the super node $X_{i,k}$ connects with all other concept nodes and the weights of their edges are the ranking scores of concepts.
The relations between $X_{i,k}$ and other concept nodes belong to one type. The initial embeddings of concept nodes in $\mathcal{G'}_{i,k}^{s}$ are pre-computed by a PLM directly without fine-tuning. Exceptionally, the initial embedding of node $X_{i,k}$ is the embedding of its corresponding dialogue context (i.e., the concatenation of $P_i$, $C_i$ and $r_{i,k}$, $\textit{input}^a_{i,k}$) via the \ac{PLM} in SinLG, which is actually the output of \textbf{Eq.(\ref{eq:lm1})}, $\boldsymbol{h}'_{i,k}$. In summary, \textbf{Trans b} includes the following operation:
\begin{equation}
	\label{eq:transb}
	\begin{split}
	    E_{i,k} & = \{ \textbf{h}_j | \textbf{h}_j = f_e (v_j), v_j \in \mathcal{V}_{i,k}^{s}\}, \\
		\mathcal{V'}_{i,k}^{s}, \mathcal{E'}_{i,k}^{s}, \mathcal{R'}_{i,k}^{s} &= \mathcal{V}_{i,k}^{s} \cup X_{i,k},  \quad \mathcal{E}_{i,k}^{s} \cup \mathcal{E}_{X_{i,k}}, \quad \mathcal{R}_{i,k}^{s} \cup \mathcal{R}_{X_{i,k}}, \\
		\mathcal{G'}_{i,k}^{s} &= (\mathcal{V'}_{i,k}^{s}, \mathcal{E'}_{i,k}^{s}, \mathcal{R'}_{i,k}^{s}), \\
    	E'_{i,k} & = E_{i,k} \cup \textbf{h}'_{i,k},\\
        \textit{input}^b_{i,k} &= (\mathcal{G'}_{i,k}^{s}, E'_{i,k}),
	\end{split}
\end{equation}
where $E'_{i,k}$ is the initial embedding set of all nodes in $\mathcal{V}'^i_{s}$, $\mathcal{E}_{X_{i,k}}$, $\mathcal{R}_{X_{i,k}}$ denotes the edges and relation types that are derived from the context node $X_{i,k}$.

\subsubsection{\textbf{Natural Language Understanding via \ac{PLM}}}\label{sec:plm}
In our framework, the PLM is utilized to encode the dialogue text which is fed forward into the \ac{PLM} after the concatenation in \textbf{Trans a}:
\begin{equation}
\label{eq:lm1}
\begin{split}
	\boldsymbol{h}'_{i,k}&=f'_e(\textit{input}^a_{i,k} ),
\end{split}
\end{equation}
\begin{equation}
	\label{eq:lm2}
	\begin{split}
		\boldsymbol{\hat{y}}_{i,k}&=f_d(\boldsymbol{h}'_{i,k}),\\
		f_d(\boldsymbol{\beta}) &= \delta (\boldsymbol{\beta} \cdot \boldsymbol{w} + b),
	\end{split}
\end{equation}
where $\boldsymbol{h}'_{i,k}$ is the first hidden state from the PLM's output, and $f_d$ is the final prediction layer which is a multi-layer perceptron where $\boldsymbol{w}$, $b$ are learnable parameters, and $\delta$ represents the activation function. Note that $\textbf{h}'_{i,k}$ here is different from the one we utilize in \textbf{Eq.(\ref{eq:score})}. The encoder in \textbf{Eq.(\ref{eq:score})} is used in zero-shot learning style while the one in \textbf{Eq.(\ref{eq:lm1})} will be fine-tuned in the subsequent training process. The rationale that we use a PLM in different styles is the directed employment of the PLM in \textbf{Eq.(\ref{eq:score})} allows all scores to be pre-computed, and then the efficiency of SinLG can be guaranteed.

\subsubsection{\textbf{Structured Knowledge Learning via \ac{GNN}}}\label{sec:gnn}
In this section, the transformation result via \textbf{Trans b}, $\textit{input}^b_{i,k} $ is fed into a GAT-based~\cite{velivckovic2017graph} graph neural network where the representation of the context node $X_{i,k}$ is learned via aggregating the messages passed from its neighbors. Taking the context node $X_{i,k}$ as an example, we formulate a single layer of propagation operation in a subgraph as:
\begin{equation}
\label{eq:gnn}
	\begin{split}
		\boldsymbol{h}^{\ell+1}_{X_{i,k}} = \delta ((\boldsymbol{h}^{\ell}_{X_{i,k}} \oplus \varphi_{agg}\{\textbf{h}_j|v_j \in \mathcal{N}_{X_{i,k}} \})\cdot \boldsymbol{w}'),
	\end{split}
\end{equation}
where $\boldsymbol{h}^{\ell}_{X_{i,k}}$ denotes the embedding of the context node $X_{i,k}$ for $\ell$-th layer, and its initial embedding $\boldsymbol{h}^{(0)}_{X_{i,k}}$ is the output of the PLM, i.e., $\textbf{h}'_{i,k}$. $\mathcal{N}_{X_{i,k}}$ represents the neighbors' set of $X_{i,k}$ and $\boldsymbol{w}'$ is a learnable parameter. $\varphi_{agg}$ is the attention aggregating function where the node type, relation, and concept score are all considered during this process~\cite{yasunaga2021qa}. Finally, we denote $\boldsymbol{h}_{X_{i,k}}$ as the final embedding of the context node $X_{i,k}$ after the message passing from subgraph $\mathcal{G'}_{i,k}^{s}$, which fuses the external commonsense knowledge into the dialogue context's representation from the PLM.

\subsection{KG-Guided Training}\label{sec:KG-guidance}
Although PLMs have achieved various impressive results on all kinds of downstream NLP tasks, they are not so trustworthy when facing complicated understanding tasks that require advanced reasoning ability based on external commonsense knowledge~\cite{liu2021graph, zhang2021greaselm}. Hence, we design the KG-guided training process in our framework to enhance the reasoning ability of PLMs with extra common sense from KG. As \textbf{Fig.\ref{fig:model-training}} depicts, both the \ac{PLM} and \ac{GNN} generate a representation vector of the context, i.e., $\boldsymbol{h}'_{i,k}$ and $\boldsymbol{h}_{X_{i,k}}$, respectively. The final predicted scores of different candidate responses are obtained via the representation vectors from the PLM. We design two sub-functions for the final loss in the following sections.

\subsubsection{\textbf{Similarity Loss.}}\label{sec:similarity}
As we all have known, PLMs have memorized all kinds of knowledge and language statistics rules during their pre-training process and exhibited extraordinary performance on many downstream tasks, which makes it impossible to ignore their strong capability. Therefore, we regard the PLM as our main backbone and assist its fine-tuning with the latent space acquired via GNN's information augmentation from external KG. Specifically, we define a similarity loss between the two representation vectors, $\boldsymbol{h}'_{i,k}$ and $\boldsymbol{h}_{X_{i,k}}$:
\begin{equation}
	\mathcal{L}_{cos} = - \frac{\boldsymbol{h}'_{i,k} \cdot \boldsymbol{h}_{X_{i,k}}}{max(||\boldsymbol{h}'_{i,k}||^2_2 \cdot ||\boldsymbol{h}_{X_{i,k}}||^2_2, \epsilon)},
	\label{eq:similarity-loss}
\end{equation}
where $\mathcal{L}_{cos}$ denotes the loss based on cosine similarity, $||.||^2_2$ represents the $\ell2$ normalization, and $\epsilon$ is a small additive term in case the denominator is 0, for instance, $10^{-8}$. In this way, the external commonsense knowledge aggregated via GNN can be transferred into the PLM to arouse its related memories and obtain a better performance on reasoning-required tasks. As an augmentation of the supervision signal, the similarity loss enables our framework to better deal with data scarcity.
\subsubsection{\textbf{Inference Loss.}}\label{sec:inference}
After the KG guidance via the similarity loss, we optimize the final performance of the PLM through the inference loss. The final predicted scores of different candidate responses have been obtained via the representation vector from the PLM according to \textbf{Eq.(\ref{eq:lm2})}. Then the inference loss is defined as follows:
\begin{equation}
	\mathcal{L}_{bce}= - (\boldsymbol{y}_{i,k}\log \boldsymbol{\hat{y}}_{i,k} + (1-\boldsymbol{y}_{i,k})\log(1-\boldsymbol{\hat{y}}_{i,k})),
	\label{eq:bce-loss}
\end{equation}
where $\mathcal{L}_{bce}$ represents the loss function based on binary cross entropy. 
\subsubsection{\textbf{Optimization Strategy.}}\label{sec:optimization}
With the definitions based on the similarity between $\boldsymbol{h}'_{i,k}$, $\boldsymbol{h}_{X_{i,k}}$ and inference results, our final loss function is formulated as:
\begin{equation}
	\mathcal{L} = \alpha \mathcal{L}_{bce} + (1-\alpha) \mathcal{L}_{cos},
	\label{eq:final-loss}
\end{equation}
where $\alpha$ refers to weighting factors, aiming to prioritize the output of a certain loss function over the other, which will be discussed in the evaluation. Above all, the model parameters are optimized by minimizing the loss defined in \textbf{Eq.(\ref{eq:final-loss})}. All parameters are updated via the Stochastic Gradient Descent (SGD) method. In specific, we utilize AdamW~\cite{loshchilov2017decoupled}, one of the SGD's variants to optimize the model parameters of SinLG.

\subsection{Efficient Inference}\label{sec:efficiency-improving}
We design the KG-guided training process via adding similarity loss instead of a direct embedding fusion (e.g., concatenating $\boldsymbol{h}_{X_{i,k}}$ from the GNN with the dialogue context embedding, $\boldsymbol{h}'_{i,k}$ from the PLM) so that we can conduct inference with only the PLM part as shown in \textbf{Fig.\ref{fig:model-inference}}. The similarity loss helps the PLM part obtain the knowledge from the GNN part and achieve better performance. Hence, it is not only an effective way of incorporating commonsense knowledge into the PLM for the MRS task, but also beneficial for inference efficiency. 
In order to utilize the commonsense knowledge from \ac{KG}, we need to conduct a bunch of tedious procedures: tokenization of the dialogue context, entity linking between context tokens and concepts in \ac{KG}, similarity calculation between concepts and the dialogue context, ranking and filtering of those concepts, and construction of subgraphs with related concepts. With our designed framework, during inference, we can not only get rid of the heavy preprocessing operations but also bypass the  generation and representation learning for any subgraphs, because the knowledge memory of the PLM is now aroused by maximizing the similarity loss. This will help increase the inference efficiency as well as reduce the computational complexity of the dialogue system.

\section{Evaluation}\label{sec:evaluation}
In this section, we showcase the advantages of SinLG on the MRS task with auxiliary information from a public commonsense knowledge graph through extensive experiments. In specific, we try to answer the following research questions:
\begin{itemize}
	\item[\textbf{RQ1}]: How is the effectiveness of SinLG on MRS task compared with existing methods?
	\item[\textbf{RQ2}]: How is the effectiveness of SinLG on MRS task under varying levels of understanding difficulty?
	\item[\textbf{RQ3}]: How is the effectiveness of SinLG on MRS task under low data availability?
	\item[\textbf{RQ4}]: How does SinLG benefit from each component of the proposed architecture?
	\item[\textbf{RQ5}]: How does the major hyper-parameter affect the performance of SinLG?
\end{itemize}

\subsection{Datasets}\label{sec:datasets}
\begin{table}
	\caption{The overall statistics of PERSONA-CHAT.}
	\vspace{-2ex}
	\small
	\renewcommand{\arraystretch}{1.2}
	\label{tab:datasets1}
	\begin{threeparttable}
	\begin{center}
		\begin{tabular}{p{2cm}<{\centering} p{1.2cm}<{\centering} p{1.2cm}<{\centering} p{1.2cm}<{\centering}}
			\toprule 
			\multirow{2}*{Category} &\multicolumn{3}{c}{PERSONA-CHAT} \\ \cline{2-4}
			& Train &Dev  &Test   \\  \hline
			{\textbf{Dialogues}}   &8,939 &1,000 &968     \\	
			{\textbf{Personas}}    &955 &100 &100         \\  
			{\textbf{mrsSamples}}  &65,719&7,801&7,512    \\   \hline
			\bottomrule
		\end{tabular}
	\begin{tablenotes}
     \scriptsize
     \item[1] The mrsSample is defined in \textbf{PROBLEM \ref{def:problem}}.
   \end{tablenotes}
  \end{center}
  \end{threeparttable}
\end{table}

\begin{table}
	\caption{The length statistics of PERSONA-CHAT variants.}
	\vspace{-2ex}
	\small
    \renewcommand{\arraystretch}{1.1}
	\label{tab:datasets2}
	\begin{threeparttable}
	\begin{center}
		\begin{tabular}{cccccccc}
			\toprule 
			\multicolumn{2}{c}{Dataset} &$Min_{len}$ &$Max_{len}$ &$Ave_{len}$ &$Len>100$ &$Len>256$ &$Len>512$ \\ \hline
			\multirow{3}*{Original}
			&Train  &35 &608 &160 &79.866\% &6.054\% &0.010\% \\
			&Dev  &51 &388 &168 &82.504\% &9.037\% &0\% \\
			&Test   &50 &406 &165 &81.373\% &7.238\% &0\% \\  \hline
			\multirow{3}*{Revised}
			&Train  &36 &603 &160 &79.939\% &5.920\% &0.010\% \\
			&Dev    &50 &394 &169 &82.981\% &9.253\% &0\% \\
			&Test   &52 &401 &166 &82.039\% &7.873\% &0\% \\  \hline
		\bottomrule
		\end{tabular}
	\begin{tablenotes}
     \scriptsize
     \item[1] $Min_{len}$, $Max_{len}$, $Ave_{len}$ represent minimum, maximum, and average length of mrsSamples in each dataset.
     \item[2] $Len>100$, $Len>256$, $Len>512$ indicate the ratios of mrsSamples with context lengths more than $100$, $256$, and $512$ among every dataset.
   \end{tablenotes}
  \end{center}
  \end{threeparttable}
\end{table}

We conduct experiments on two variants of PERSONA-CHAT~\cite{zhang2018personalizing}, that is, the dialogues with original personas and its revised version whose detail statistics are listed in \textbf{Table \ref{tab:datasets1}} and \textbf{Table \ref{tab:datasets2}}. The revised version is rephrased from the original one in an implicit way which is more challenging than the original one in the understanding complexity so that we can testify the effectiveness of our model under different levels. We choose ConceptNet~\cite{speer2017conceptnet} as our auxiliary knowledge graph, which is a widely-used~\cite{jain2022word, zhou2022eventbert, yasunaga2021qa, huang2020enhanced, sap2019atomic, bosselut2019comet} commonsense source. 

As \textbf{Table \ref{tab:datasets1}} shows, PERSONA-CHAT possesses $10,907$ complete conversations, $8939$ for training, $1,000$ for development, and $968$ for test, where dialogues have $6\sim8$ conversation turns. Response selection is conducted at each round of a complete conversation, which leads to $81,032$ mrsSamples (which is defined in the definition of PROBLEM \ref{def:problem}.) in total, $65,719$ for training, $7,801$ for development, and $7,512$ for testing. The response candidate set of each turn includes 20 choices, where a $1$ positive response is the true one from human beings while $19$ negative ones are stochastically selected. As shown in \textbf{Fig.\ref{fig:dialogue-ex}}, the revised version has the same dialogue context as the original one but its personas are rephrased, generalized, or specialized, which no doubt increases the difficulty of the MRS task. We conduct statistics about the context lengths of mrsSamples where $Min_{len}$, $Max_{len}$, $Ave_{len}$ represent the minimum, maximum, and average lengths of each dataset and the last three columns are the ratios of mrsSamples with context lengths more than $100$, $256$, and $512$ among every dataset. As \textbf{Table \ref{tab:datasets2}} shows, the context lengths of training datasets vary more, ranging from $30+$ to $600+$ while those in development and test datasets range from $50+$ to $400\pm$. This could increase the robustness of the model after training on these training datasets but the development and test datasets seem not that comprehensive because they leave out the validating chance of extreme cases. Moreover, we can observe that samples with lengths between $Min_{len}$ and $100$ take up the maximum proportion and the second category is $Len>100$, then $Len>256$, $Len>512$, orderly. These statistics are good references for the experiment setting, i.e., the maximum sequence lengths of PLMs, which have a significant impact on their performance. For instance, if the hardware resources are limited, the max sequence length can be chosen between 100 and 256 to test uniformly.

ConceptNet is a public multilingual graph of commonsense knowledge where labeled edges connect phrases and words. With the combination of several different sources (i.e., Facts acquired from Open Mind common sense~\cite{singh2002public}, a subset of DBpedia~\cite{auer2007dbpedia}, JMDict~\cite{breen2004jmdict}, Open Multilingual WordNet~\cite{bond2013linking}, etc.), ConceptNet possesses over $21$ million edges and over $8$ million nodes, whose English vocabulary consists of about $1500,000$ nodes. 

\subsection{Baselines}\label{sec:baselines}
To evaluate the effectiveness of SinLG on MRS, we make comparisons with state-of-the-art results from~\cite{gu2020filtering, gu2021partner}. Their brief introductions are listed as follows:

\begin{itemize}
	\item{\textbf{Starspace:}} Starspace~\cite{wu2018starspace} learns entity embeddings separately and scores the response candidates by computing the similarity between them and the conversation.
	
	\item{\textbf{PM:}} Ranking Profile Memory network (PM) is proposed in the publishing paper of PERSONA-CHAT,~\cite{zhang2018personalizing} which uses an attention mechanism to identify the relevant lines of the profile with the dialogue context.
	
	\item{\textbf{KV-PM:}} As an improvement of PM, the Key-Value Profile Memory network (KV-PM) is put forward~\cite{miller2016key}. It calculates attention scores over keys and obtains the values instead of using the same keys as the original one, which can exceed the profile memory network according to the definition of key-value pairs on specific tasks.
	
	\item{\textbf{Transformer:}} Transformer~\cite{vaswani2017attention} is an encoder-decoder memory network based only on the attention mechanism. It achieves state-of-the-art performance on the next utterance retrieval task under multi-turn dialogues. The experiment of the transformer employs the setting in~\cite{mazare2018training}, which only utilizes the encoding architecture. 
	
	\item{\textbf{DGMN:}} Document-Grounded Matching Network (DGMN)~\cite{zhao2019document} is one of the several pioneering studies that incorporate extra information (a related document) into the context besides the dialogue text itself. DGMN constructs rich representation for the context and document via self-attention to do matching for the dialogue and candidate response better.
		
	\item{\textbf{DIM:}} Dually Interactive Matching network (DIM)~\cite{gu2019dually} ranks response candidates via interactive matching for both response-context pairs and response-persona pairs respectively.
	
	\item{\textbf{FIRE:}} FIRE~\citet{gu2020filtering} identifies relative knowledge for context bidirectionally before matching response candidates with the conversation context and corresponding knowledge.
	
	\item{\textbf{BERT:}} BERT~\cite{devlin2019bert} is developed by Google, which is based on the encoder module of the transformer, targeting the pre-training of natural language processing (NLP). It exhibited state-of-the-art performance on many natural language understanding tasks, including GLUE, SQuAD, and SWAG, regarded as the SOTA baseline.
	
	\item{\textbf{Roberta:}} According to proposing a set of important design choices and training strategies, and introducing alternatives, Roberta~\cite{liu2019roberta} achieves better downstream task performance than BERT. Roberta employed a novel dataset, CCNEWS, which testifies that utilizing more data for pre-training further boosts performance on downstream tasks.
\end{itemize}

\subsection{Experiment Setting}\label{sec:expset}
We evaluate the model performance with the identical metrics $R_{n}@k$ and MRR which are commonly used on the MRS task~\cite{tao2020improving,gu2021partner,whang2021response}. The equations of them are as follows:
\begin{equation}
\label{eq:rmse}
R_n@k = \frac{1}{|D|} \sum_{i=1}^{|D|}f(i), \quad f(i)=\left\{
\begin{aligned}
1,\quad rank_i \leqslant k, \\
0,\quad rank_i > k.\\
\end{aligned}
\right.\\
\end{equation}

\begin{equation}
\label{eq:smape}
MRR = \frac{1}{|D|} \sum_{i=1}^{|D|} \frac{1}{rank_i},
\end{equation}
where $|D|$ represents the total number of samples in the target dataset and $rank_i$ denotes the rank of the ground-truth response in the $i$ sample. As there are $20$ response choices in PERSONA-CHAT, we choose $R_{20}@1$, $R_{20}@2$, $R_{20}@5$ and MRR as our final metrics. 

The data splitting of PERSONA-CHAT is fixed, which is shown in \textbf{Table \ref{tab:datasets1}}. According to the statistics of datasets in \textbf{Table \ref{tab:datasets2}}, the default max sequence length of \ac{PLM} is set as 512 which will cover nearly all samples' lengths. The default number of nodes, embedding size, and the number of layers number in the \ac{GNN} is set as 200, 200, and 5. For the sake of fairness, we set the batch size of all models as 64 and use the learning rate that literature~\cite{yasunaga2021qa} recommends. The loss weight $\alpha$ is set as 0.5 by default. We implement our solutions with Pytorch 1.9 and Python 3.8.


\subsection{Performance Analysis}\label{sec:performance}
\subsubsection{\textbf{The Effectiveness of SinLG}}\label{sec:effectiveness} In order to evaluate the effectiveness of SinLG, we conduct extensive experiments on the two variants of PERSONA-CHAT original and revised. We will analyze its effectiveness from three perspectives: comparisons with existing methods, comparisons under varying levels of understanding difficulty, and comparisons under the low-resource scenario.

\begin{table*}
	\caption{Results of different models on PERSONA-CHAT.}
	\vspace{-2ex}
	\small
	\renewcommand{\arraystretch}{1.2}
	\label{tab:effectiveness}
	\begin{threeparttable}
	\begin{tabular}{ccccccccc}
		\Xhline{1pt}
		\multirow{2}*{Model}  
		                                      & \multicolumn{4}{c}{Original}   &\multicolumn{4}{c}{Revised} \\ \cline{2-9} 
		                                      & $R_{20}@1$& $R_{20}@2$ & $R_{20}@5$ &MRR   & $R_{20}@1$& $R_{20}@2$ & $R_{20}@5$ &MRR \\  \Xhline{1pt}
		Starspace~\cite{wu2018starspace}    &49.1&60.2&76.5 &- &32.2&48.3&66.7 &-\\
		PM~\cite{zhang2018personalizing} &50.9 &60.7 &75.7 &- &35.4 &48.3 &67.5 &-\\
		KV-PM~\cite{miller2016key}  &51.1 &61.8 &77.4 &- &35.1 &45.7 &66.3 &-\\
		Transformer~\cite{mazare2018training} &54.2 &68.3 &83.8 &- &42.1 &56.5 &75.0 &-\\
		DGMN~\cite{zhao2019document} &67.6 &80.2 &92.9 &- &58.8 &62.5 &87.7 &- \\
		DIM~\cite{gu2019dually} &78.8 &89.5 &97.0 &-  &70.7 &84.2 &95.0 &- \\
		FIRE~\cite{gu2020filtering} &81.6 &91.2 &97.8 &- &74.8 &86.9 &95.9 &- \\
		BERT~\cite{devlin2019bert} &85.49 &94.0 &98.52 &91.27 &79.41 &90.27 &97.5 &87.25 \\
		Roberta~\cite{liu2019roberta} &85.22	&94.36	&98.67	&91.19 &80.62	&90.52	&97.66	&87.94 \\
		SinLG (ours) &\textbf{86.91} &\textbf{94.61} &\textbf{98.91} &\textbf{92.16} &\textbf{82.59} &\textbf{91.97} &\textbf{97.94} &\textbf{89.29} \\ \Xhline{1pt}
	\end{tabular}
	\begin{tablenotes}
     \scriptsize
     \item[1] The experimental results of compared models under $R_{20}@1$, $R_{20}@2$, $R_{20}@5$ and MRR is on the \textbf{test} datasets of PERSONA-CHAT original and revised.
     \item[2] The results of Starspace, PM, KV-PM, Transformer, DGMN, DIM, and FIRE are from literature~\cite{gu2020filtering, gu2021partner}.
     \item[3] Note that the results of BERT, Roberta, and SinLG (based on Roberta) is the best ones under the provided default settings.
     \item[4] Numbers in boldface are the best results for corresponding metrics with p-value < 0.05.
   \end{tablenotes}
  \end{threeparttable}
\end{table*}

\begin{table*}
	\caption{Results of PLMs and SinLG with different max sequence lengths on PERSONA-CHAT original.}
	\vspace{-2ex}
	\small
	\renewcommand{\arraystretch}{1.2}
	\label{tab:msl1}
	\begin{threeparttable}
	\begin{tabular}{ccccccccc}
		\Xhline{1pt}
		\multirow{3}*{Model} 
		&\multicolumn{8}{c}{Original}\\ \cline{2-9} 
		&\multicolumn{4}{c}{Dev} &\multicolumn{4}{c}{Test} \\ \cline{2-9} 
		&$R_{20}@1$ &$R_{20}@2$ &$R_{20}@5$ &MRR &$R_{20}@1$ &$R_{20}@2$ &$R_{20}@5$ &MRR \\  \Xhline{1pt}
		\rowcolor{gray!15}BERT-256 &80.94 &88.75 &94.03 &86.97 &81.16 &89.63 &94.94 &87.46 \\
		\rowcolor{gray!15}SinLG-bert-256  &82.19 &91.31 &97.31 &88.8 &81.67 &91.59 &97.83 &88.71 \\
		Roberta-256  &80.73 &87.89 &93.12 &86.45 &81.32 &88.91 &94.57 &87.27 \\
		SinLG-roberta-256 &85.39 &92.71 &97.67 &90.75  &84.28 &92.53 &97.88 &90.2 \\
		\rowcolor{gray!15}BERT-512  &86.04 &93.87 &98.23 &91.45 &85.49 &94.0 &98.52 &91.27  \\
		\rowcolor{gray!15}SinLG-bert-512 &85.04 &93.73 &98.3 &90.93  &84.57 &93.89 &98.67 &90.78 \\
		Roberta-512 &86.23 &93.76 &98.13 &91.49  &85.22 &94.36 &98.67 &91.19 \\
		SinLG-roberta-512 &87.64 &94.48 &98.77 &92.47  &86.91 &94.61 &98.91 &92.16 \\
         \Xhline{1pt}
	\end{tabular}
	\begin{tablenotes}
     \scriptsize
     \item[1] Model$-256/512$ means the maximum sequence length of the context is set as 256 or 512.
   \end{tablenotes}
   \end{threeparttable}
\end{table*}

\begin{table*}
	\caption{Results of PLMs and SinLG with different max sequence lengths on PERSONA-CHAT revised.}
	\vspace{-2ex}
	\small
	\renewcommand{\arraystretch}{1.2}
	\label{tab:msl2}
	\begin{threeparttable}
	\begin{tabular}{ccccccccc}
		\Xhline{1pt}
		\multirow{3}*{Model} 
		&\multicolumn{8}{c}{Revised}\\ \cline{2-9} 
		&\multicolumn{4}{c}{Dev} &\multicolumn{4}{c}{Test} \\ \cline{2-9} 
		&$R_{20}@1$ &$R_{20}@2$ &$R_{20}@5$ &MRR &$R_{20}@1$ &$R_{20}@2$ &$R_{20}@5$ &MRR \\  \Xhline{1pt}
		\rowcolor{gray!15}BERT-256  &75.84 &85.82 &93.05 &83.59  &75.29 &86.02 &93.54 &83.47  \\
		\rowcolor{gray!15}SinLG-bert-256  &77.95 &88.73 &96.42 &85.98  &78.0 &88.97  &96.82 &86.13 \\
		Roberta-256  &75.99 &85.28 &92.94 &83.39 &76.44 &85.74  &92.94 &83.87   \\
		SinLG-roberta-256  &80.12 &89.77 &96.67 &87.31 &78.65 &89.46  &96.98 &86.57   \\
		\rowcolor{gray!15}BERT-512  &79.98 &90.5 &97.69 &87.6 &79.41 &90.27 &97.5 &87.25  \\
		\rowcolor{gray!15}SinLG-bert-512  &80.03 &90.48 &97.54 &87.63  &80.06 &90.84  &97.71 &87.69   \\
		Roberta-512  &81.43 &90.86 &97.32 &88.36  &80.62 &90.52  &97.66 &87.94   \\
		SinLG-roberta-512  &83.36 &92.17 &97.97 &89.72  &82.59 &91.97 &97.94 &89.29   \\
         \Xhline{1pt}
	\end{tabular}
	\begin{tablenotes}
     \scriptsize
     \item[1] Model$-256/512$ means the maximum sequence length of the context is set as 256 or 512.
   \end{tablenotes}
   \end{threeparttable}
\end{table*}

\paragraph{\textbf{Comparisons with existing methods (RQ1).}}
\textbf{Table \ref{tab:effectiveness}} summarize the results of all compared methods under four metrics on the test datasets of PERSONA-CHAT original and revised. Note that results of the former seven models (i.e., Starspace, PM, KV-PM, Transformer, DGMN, DIM, FIRE) are referred to the literature~\cite{gu2020filtering, gu2021partner}. The models in \textbf{Table \ref{tab:effectiveness}} are in chronological order from top to bottom. We make the following observations:
\begin{itemize}
	\item Research on MRS achieves impressive progress from simple unidirectional encoders (i.e., Starspace, PM, KV-PM, Transformer) to interactional and bidirectional ones (i.e., DGMN, DIM, FIRE, BERT, Roberta, SinLG). The performance gains from Transformer to DGMN and then DIM are remarkable, which means that interactional fusions and bidirectional semantic distinguishing are very significant operations for natural language understanding.
	\item Previous works employ all kinds of detailed mechanisms to refine their models. Then PLMs show absolute predominance because of their bidirectional attention architecture and huge parameters, defeating all the previous models and completing the MRS task with good results. 
	\item The performance gain of our SinLG from PLMs shows that the external commonsense knowledge could provide auxiliary information bias to improve the performance of PLMs. This indicates that PLMs do not always exhibit their best performance via fine-tuning, and extra knowledge can assist to arouse their related memory for some specific tasks. 
\end{itemize}
As a result, our method can achieve state-of-the-art results on both variants of PERSONA-CHAT with reasonable superiority. In order to illustrate the effectiveness of SinLG more specifically, we provide \textbf{Table \ref{tab:msl1}} and \textbf{Table \ref{tab:msl2}} where we compare PLM base models with SinLG under different max sequence lengths. From the results, we can see that SinLG outperforms base models steadily, especially on PERSONA-CHAT revised or when the max sequence length is 256.

\begin{figure*}[tb!]
	\centering
	\setlength{\abovecaptionskip}{0.15cm}
	\subfigure[$R_{20}@1$]{
		\label{fig:eva1:sub1}
		\includegraphics[width=0.4\textwidth]{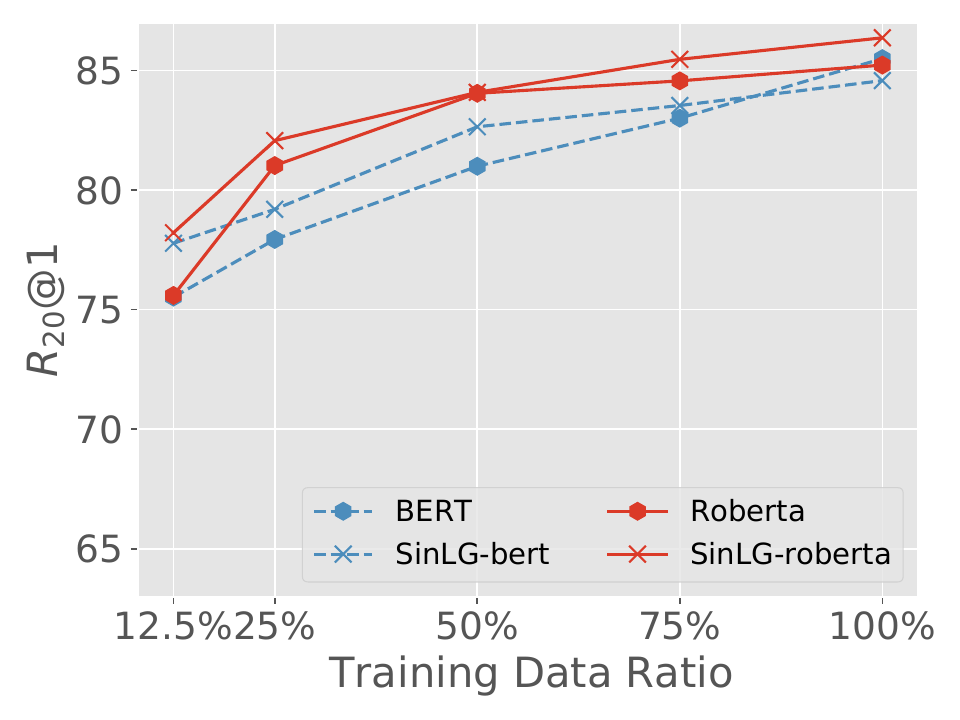}}
	\hspace{1ex}
	\subfigure[$R_{20}@2$]{
		\label{fig:eva1:sub2}
		\includegraphics[width=0.4\textwidth]{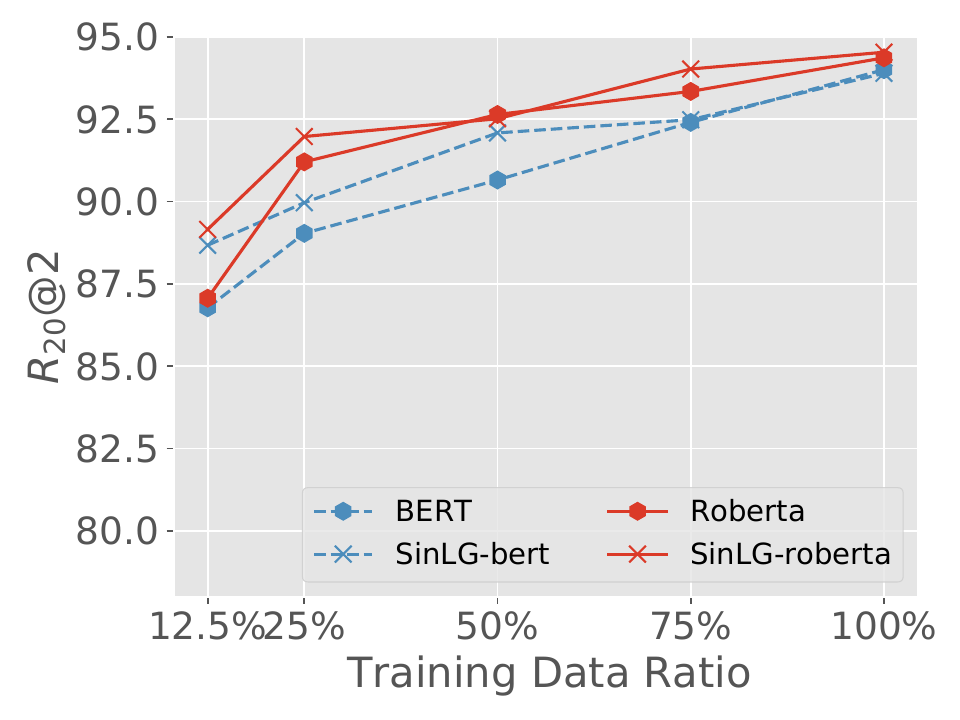}}
	\hspace{1ex}
	\subfigure[$R_{20}@5$]{
		\label{fig:eva1:sub3}
		\includegraphics[width=0.4\textwidth]{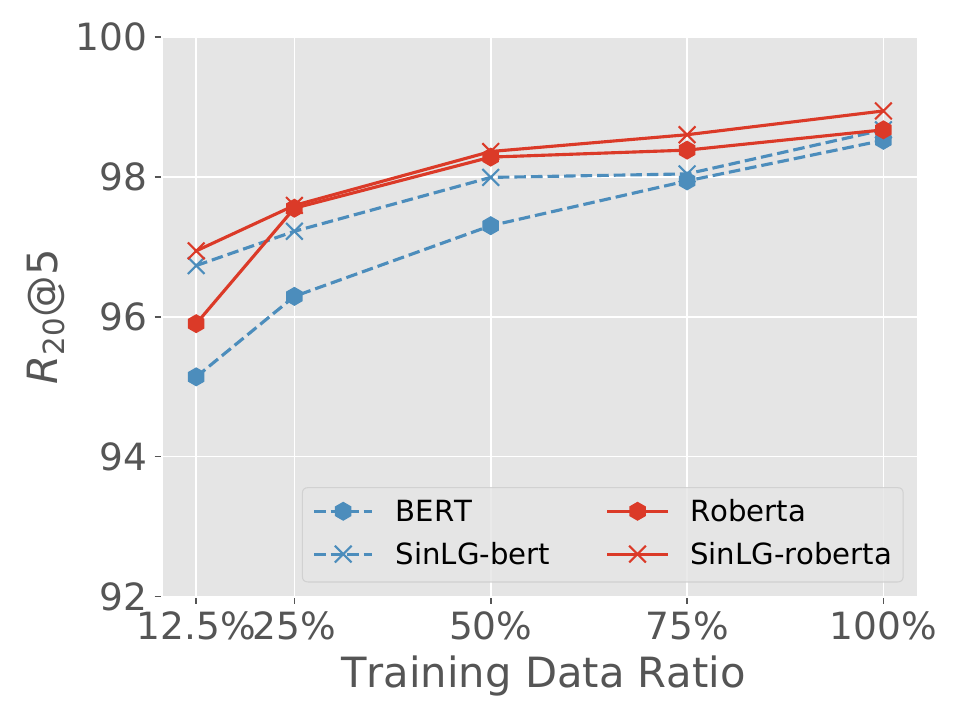}}
	\hspace{1ex}
	\subfigure[MRR]{
		\label{fig:eva1:sub4}
		\includegraphics[width=0.4\textwidth]{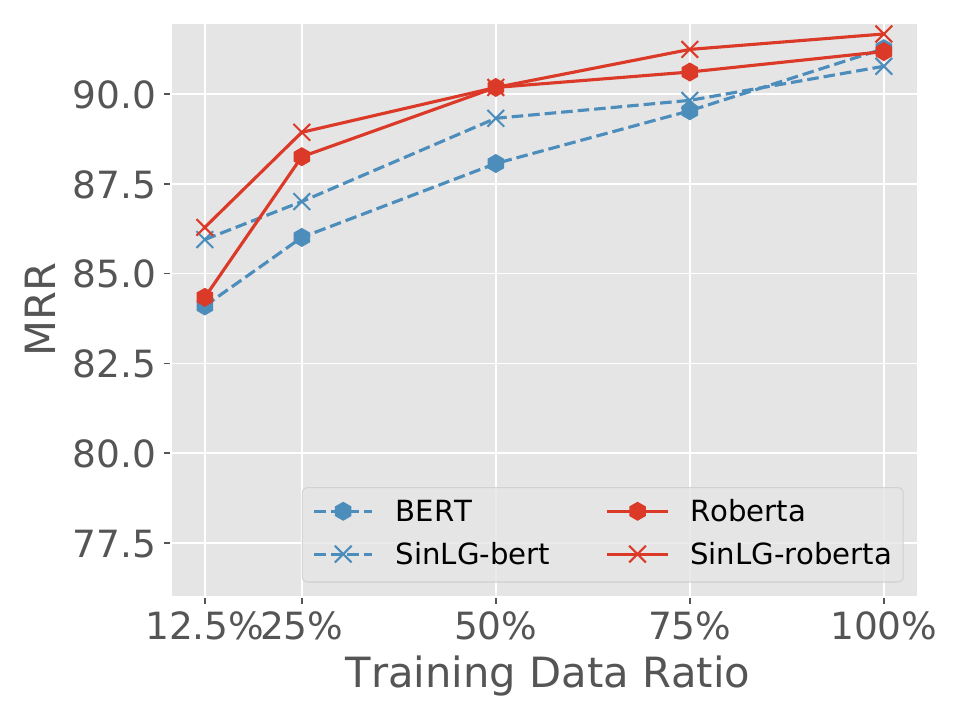}}
	\caption{SinLG performance under the low-resource scenario on PERSONA-CHAT original.}
	\label{fig:eva1} 
\end{figure*}

\begin{figure*}[tb!]
	\centering
	\setlength{\abovecaptionskip}{0.15cm}
	\subfigure[$R_{20}@1$]{
		\label{fig:eva2:sub1}
		\includegraphics[width=0.4\textwidth]{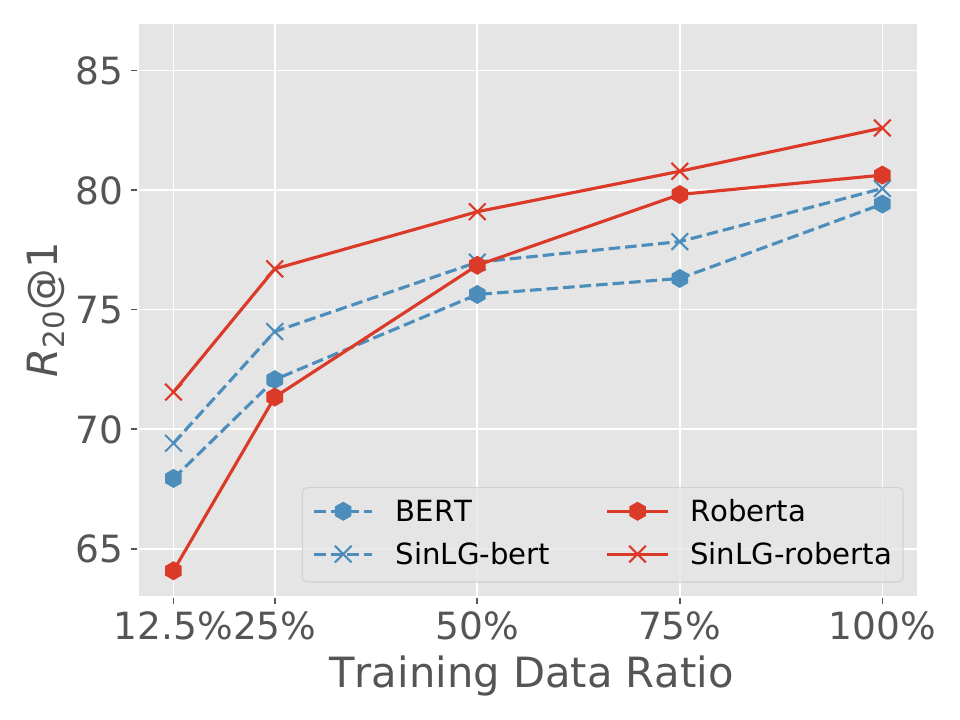}}
	\hspace{1ex}
	\subfigure[$R_{20}@2$]{
		\label{fig:eva2:sub2}
		\includegraphics[width=0.4\textwidth]{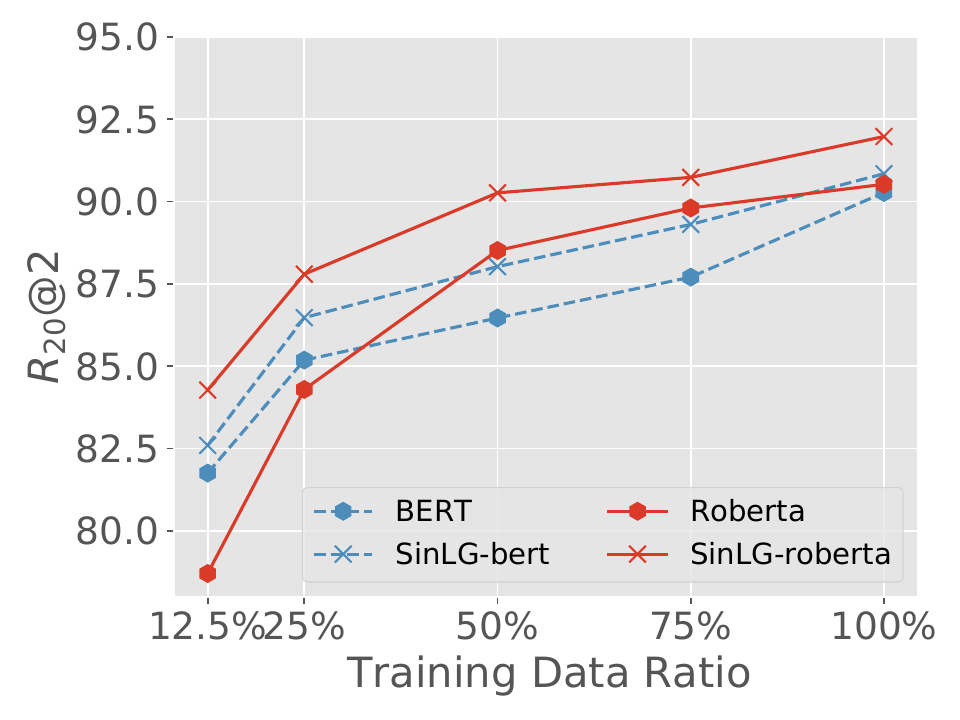}}
	\hspace{1ex}
	\subfigure[$R_{20}@5$]{
		\label{fig:eva2:sub3}
		\includegraphics[width=0.4\textwidth]{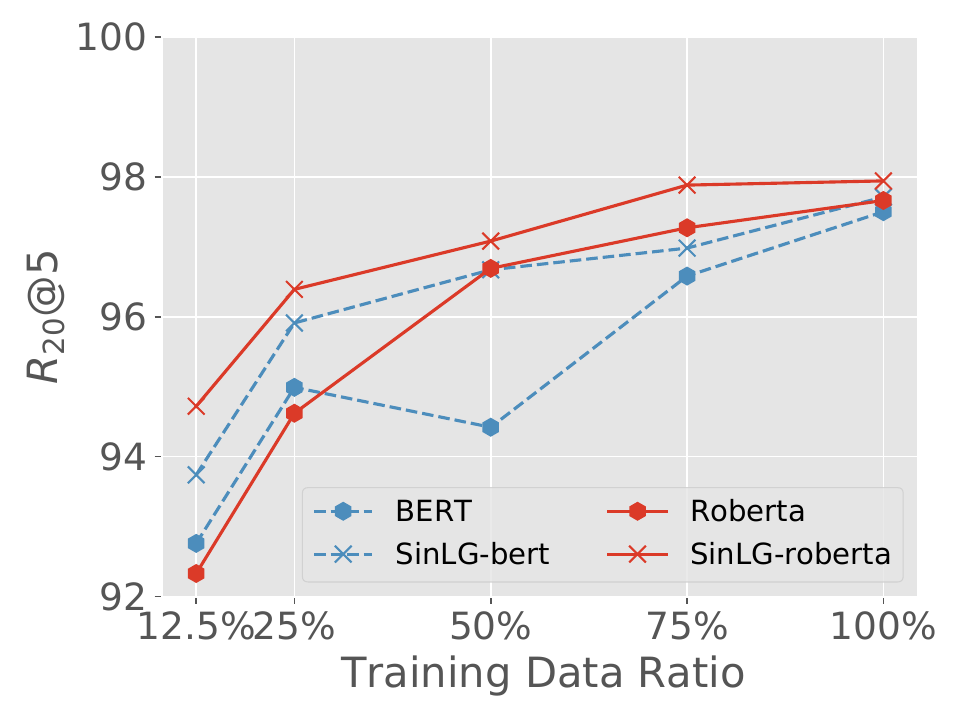}}
	\hspace{1ex}
	\subfigure[MRR]{
		\label{fig:eva2:sub4}
		\includegraphics[width=0.4\textwidth]{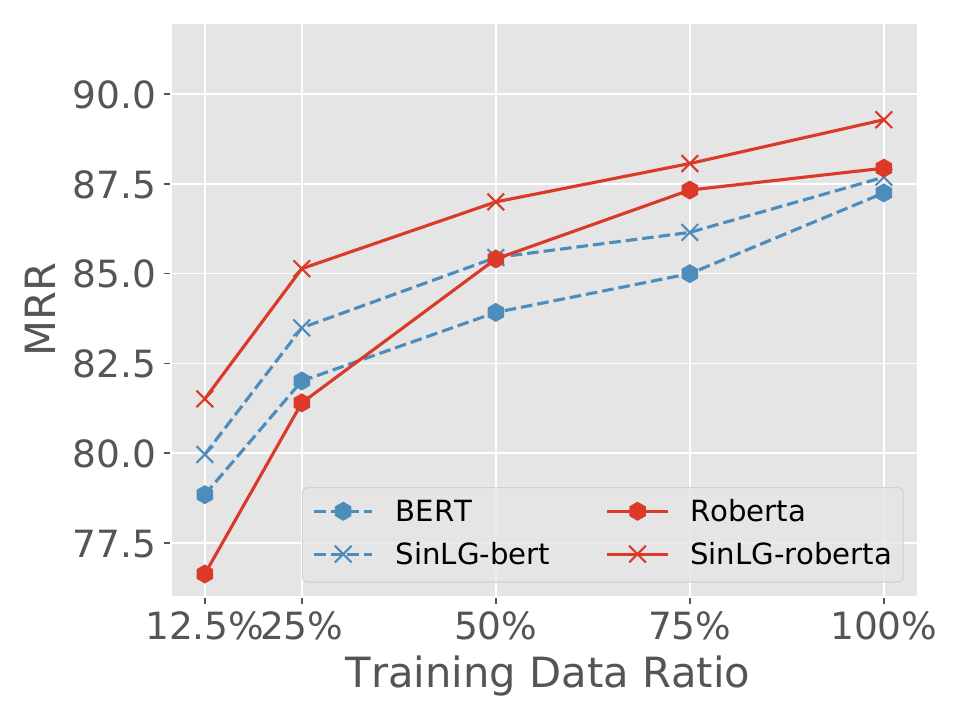}}
	\caption{SinLG performance under the low-resource scenario on PERSONA-CHAT revised.}
	\label{fig:eva2} 
\end{figure*}

\paragraph{\textbf{Comparisons under varying levels of understanding difficulty (RQ2).}}
In the above perspective, we compare our model with existing methods. In this part, we will analyze the effectiveness of our model on the two variants of PERSONA-CHAT, which are under varying levels of understanding difficulty. As aforementioned in \textbf{Section \ref{sec:datasets}}, PERSONA-CHAT revised is more difficult than PERSONA-CHAT original. Its persona descriptions are written in a more obscure way, which makes it need more complicated common sense to understand. 
\begin{itemize}
	\item From \textbf{Table \ref{tab:effectiveness}}, we can observe that our proposed model SinLG achieves more performance promotion on PERSONA-CHAT revised than PERSONA-CHAT original, and the same goes for all the other models. These improvements indicate that our proposed model could better handle more complicated understanding tasks. Hence, the difference value of Starspace's results on PERSONA-CHAT original and revised is up to 16.9 at $R_{20}@1$ while it is only 4.35 with SinLG.
	\item From \textbf{Table \ref{tab:msl1}} and \textbf{Table \ref{tab:msl2}}, it can be observed that BERT-based SinLG achieves more performance gain compared with BERT on PERSONA-CHAT revised than PERSONA-CHAT original under both experimental setting with the maximum sequence length of 256 and 512. This phenomenon illustrates that BERT can well handle simple datasets via fine-tuning while under the more challenging understanding scenario, extra common sense would help it to perform better. However, the results of Roberta-based SinLG show even performance gains compared with Roberta on both PERSONA-CHAT original and revised, which indicates that Roberta needs extra common sense to arouse its related memory even on the simple dataset. Fortunately, this auxiliary operation can assist Roberta to achieve better results than BERT, which means the extra knowledge can help Roberta more. 
	\item Comparing the same metrics in \textbf{Fig.\ref{fig:eva1}} and \textbf{Fig.\ref{fig:eva2}}, it is easy to find that the performance gain of our model on PERSONA-CHAT revised is more than on PERSONA-CHAT original under both full and low data resource scenarios, which testifies the effectiveness of our model on the more difficult understanding task again.
\end{itemize}
Overall, SinLG can obtain more promotion on PERSONA-CHAT revised than PERSONA-CHAT original. Nevertheless, in some specific cases, the performance gain of SinLG compared with its base models depends on the various sensitivity of base models to datasets and incorporating knowledge. 
In order to explore the effectiveness of SinLG more deeply, we identify such examples and perform comparisons upon cases that need external common sense and cases that do not need it. Specifically, we filter out samples whose score sums of related concepts from KG are within or out of the top 10\% to testify the performance of SinLG. Then we find that the performance gain of our model from KG is more outstanding on the top 10\% samples and trivial on the rest samples. The results are shown in \textbf{Table \ref{tab:effectiveness1}} and \textbf{Table \ref{tab:effectiveness2}}.

\begin{table*}
	\caption{Results of PLMs and SinLG on two subsets of PERSONA-CHAT original.}
	\vspace{-2ex}
	\small
	\renewcommand{\arraystretch}{1.2}
	\label{tab:effectiveness1}
	\begin{threeparttable}
	\begin{tabular}{ccccccccc}
		\Xhline{1pt}
		\multirow{3}*{Model} 
		&\multicolumn{8}{c}{Original}\\ \cline{2-9} 
		&\multicolumn{4}{c}{Top 10\%} &\multicolumn{4}{c}{Remaining} \\ \cline{2-9} 
		&$R_{20}@1$ &$R_{20}@2$ &$R_{20}@5$ &MRR &$R_{20}@1$ &$R_{20}@2$ &$R_{20}@5$ &MRR \\  \Xhline{1pt}
		BERT  &84.69 &93.74	&99.47 &91.04 &86.88 &94.57	&99.38 &92.26 \\
            \rowcolor{gray!15}SinLG-bert &87.08 &95.07 &99.33 &92.41 &86.75 &94.71 &99.42	&92.27 \\
		Roberta &84.02 &93.16 &99 &91.1 &87.74 &95.54 &99.52 &92.7 \\
		\rowcolor{gray!15}SinLG-roberta  &86.42 &96.01 &99.6 &92.24 &88.31 &95.8 &99.11 &93.01 \\ \Xhline{1pt}
	\end{tabular}
	\begin{tablenotes}
     \scriptsize
     \item[1] The maximum sequence length of the context is set as 512 in this comparison.
     \item[2] SinLG-bert/roberta represents the PLM part of our model that employs BERT or Roberta.
   \end{tablenotes}
  \end{threeparttable}
\end{table*}

\begin{table*}
	\caption{Results of PLMs and SinLG on two subsets of PERSONA-CHAT revised.}
	\vspace{-2ex}
	\small
	\renewcommand{\arraystretch}{1.2}
	\label{tab:effectiveness2}
	\begin{threeparttable}
	\begin{tabular}{ccccccccc}
		\Xhline{1pt}
		\multirow{3}*{Model} 
		&\multicolumn{8}{c}{Revised}\\ \cline{2-9} 
		&\multicolumn{4}{c}{Top 10\%} &\multicolumn{4}{c}{Remaining} \\ \cline{2-9} 
		&$R_{20}@1$ &$R_{20}@2$ &$R_{20}@5$ &MRR &$R_{20}@1$ &$R_{20}@2$ &$R_{20}@5$ &MRR \\  \Xhline{1pt}
		BERT  &78.96 &91.48	&98	&87.32 &80.77 &91.45 &97.81	&88.1 \\
            \rowcolor{gray!15}SinLG-bert &81.09	&92.68	&98.8	&88.8 &80.82 &91.75	&97.98	&88.43 \\
		Roberta &77.63	&91.32	&97.89	&87 &81.03	&91.39	&98.4	&89.06 \\
		\rowcolor{gray!15}SinLG-roberta  &80.56	&92.68	&98.93	&88.62 &81.97	&92.6	&98.78	&89.17 \\ \Xhline{1pt}
	\end{tabular}
	\begin{tablenotes}
     \scriptsize
     \item[1] The maximum sequence length of the context is set as 512 in this comparison.
     \item[2] SinLG-bert/roberta represents the PLM part of our model that employs BERT or Roberta.
   \end{tablenotes}
  \end{threeparttable}
\end{table*}

\paragraph{\textbf{Comparisons under low data availability (RQ3).}} 
In this section, we will discuss the effectiveness of SinLG under low data availability, i.e., the shortage of the training corpus. The limited training data will bring unexpected performance loss for PLMs. Hence, we split the training set into several subsets, i.e., $12.5\%$, $25\%$, $50\%$, and $75\%$ of the whole quantity to examine the performance of each method. We keep the development and test sets unchanged in these experiments. Their results are illustrated with \textbf{Fig.\ref{fig:eva1}} and \textbf{Fig.\ref{fig:eva2}} and the ranges of the same metrics on PERSONA-CHAT original and revised are set to the same for convenient comparisons. 

\begin{itemize}
         \item From \textbf{Fig.\ref{fig:eva1}}, we can see that our model SinLG achieves an obvious performance gain with both PLM base models on PERSONA-CHAT original, especially when only $12.5\%$ and $25\%$ of the training data are provided. With the increase of training data, the performance of SinLG and the base models tend to be closer. 
         \item As \textbf{Fig.\ref{fig:eva2}} displays, the performance gain of SinLG compared with the base models is more outstanding on PERSONA-CHAT revised, especially that between SinLG and Roberta. This indicates that our proposed method has more advantages when low-data availability and more difficultly-understanding datasets come together.
         
\end{itemize}
According to the above results and analysis, we can conclude that our proposed method could achieve superior performance  when it is facing data scarcity.

\subsubsection{\textbf{Ablation Study (RQ4).}}\label{sec:ablation}
\begin{table*}
	\caption{Ablation study results on PERSONA-CHAT original.}
	\vspace{-2ex}
	\small
	\renewcommand{\arraystretch}{1.2}
	\label{tab:as1}
	\begin{threeparttable}
	\begin{tabular}{cccccccccc}
		\Xhline{1pt}
	    \multirow{2}*{Variant}&\multirow{2}*{PLM-base} &\multirow{2}*{KG}  &\multirow{2}*{GNN} &\multirow{2}*{SL} &\multirow{2}*{QO-free} &\multicolumn{4}{c}{Original-Test} \\ \cline{7-10} 
		& & & & & &$R_{20}@1$ &$R_{20}@2$ &$R_{20}@5$ &MRR \\  \Xhline{1pt}
	    Roberta  &$\surd$& & & &$\surd$ &83.56 &92.73 &98.12 &89.94  \\
	    SinLG-S0 &$\surd$&$\surd$ & & & &83.67	&92.28	&97.62	&89.79 \\
		SinLG-S1 &$\surd$&$\surd$ & & &  &86.16 &94.36 &98.95 &91.73 \\
		SinLG-S2 &$\surd$&$\surd$ & &$\surd$ &$\surd$ &85.8 &94.08 &98.8 &91.47  \\
		SinLG-S3 &$\surd$&$\surd$ &$\surd$ & & &86.31 &94.26 &\textbf{99.86} &91.8  \\
		SinLG-roberta    &$\surd$&$\surd$ &$\surd$ &$\surd$ &$\surd$ &\textbf{86.91} &\textbf{94.61} &98.91 &\textbf{92.16}  \\
         \Xhline{1pt}
	\end{tabular}
	\begin{tablenotes}
     \scriptsize
     \item[1] Checked KG indicates the model employs knowledge from the knowledge graph.
     \item[2] Checked SL means the model utilizes similarity loss to assist the training process of PLMs. Consequently, the QO-free item is checked, too, that is to say, there is no need to query entities from KG during the inference.
     \item[3] Numbers in boldface are the best results for corresponding metrics.
   \end{tablenotes}
   \end{threeparttable}
\end{table*}

\begin{table*}
	\caption{Ablation study results on PERSONA-CHAT revised.}
	\vspace{-2ex}
	\small
	\renewcommand{\arraystretch}{1.2}
	\label{tab:as2}
	\begin{threeparttable}
	\begin{tabular}{cccccccccc}
		\Xhline{1pt}
	    \multirow{2}*{Variant}&\multirow{2}*{PLM-base} &\multirow{2}*{KG}  &\multirow{2}*{GNN} &\multirow{2}*{SL} &\multirow{2}*{QO-free} &\multicolumn{4}{c}{Revised-Test} \\ \cline{7-10} 
		& & & & & &$R_{20}@1$ &$R_{20}@2$ &$R_{20}@5$ &MRR \\  \Xhline{1pt}
		Roberta  &$\surd$& & & &$\surd$ &80.25 &90.83 &97.42 &87.86  \\
		SinLG-S0 &$\surd$&$\surd$ & & & &77.85	&89.12	&96.63	&86.07 \\
		SinLG-S1 &$\surd$&$\surd$ & & & &81.82 &91.75 &97.82 &88.87 \\
		SinLG-S2 &$\surd$&$\surd$ & &$\surd$ &$\surd$ &81.7 &91.48 &97.84 &88.83  \\
		SinLG-S3 &$\surd$&$\surd$ &$\surd$ & & &81.98 &91.69 &\textbf{97.99} &88.95  \\
		SinLG-roberta    &$\surd$&$\surd$ &$\surd$ &$\surd$ &$\surd$ &\textbf{82.59} &\textbf{91.97} &97.94 &\textbf{89.29}  \\
         \Xhline{1pt}
	\end{tabular}
	\begin{tablenotes}
     \scriptsize
     \item[1] Checked KG indicates the model employs knowledge from the knowledge graph.
     \item[2] Checked SL means the model utilizes similarity loss to assist the training process of PLMs. Consequently, the QO-free item is checked, too, that is to say, there is no need to query entities from KG during the inference.
     \item[3] Numbers in boldface are the best results for corresponding metrics.
   \end{tablenotes}
   \end{threeparttable}
\end{table*}

\begin{figure*}[tb!]
	\centering
	\setlength{\abovecaptionskip}{0.15cm}
	\subfigure[SinLG-S3]{
		\label{fig:eva3:sub1}
		\includegraphics[width=0.4\textwidth]{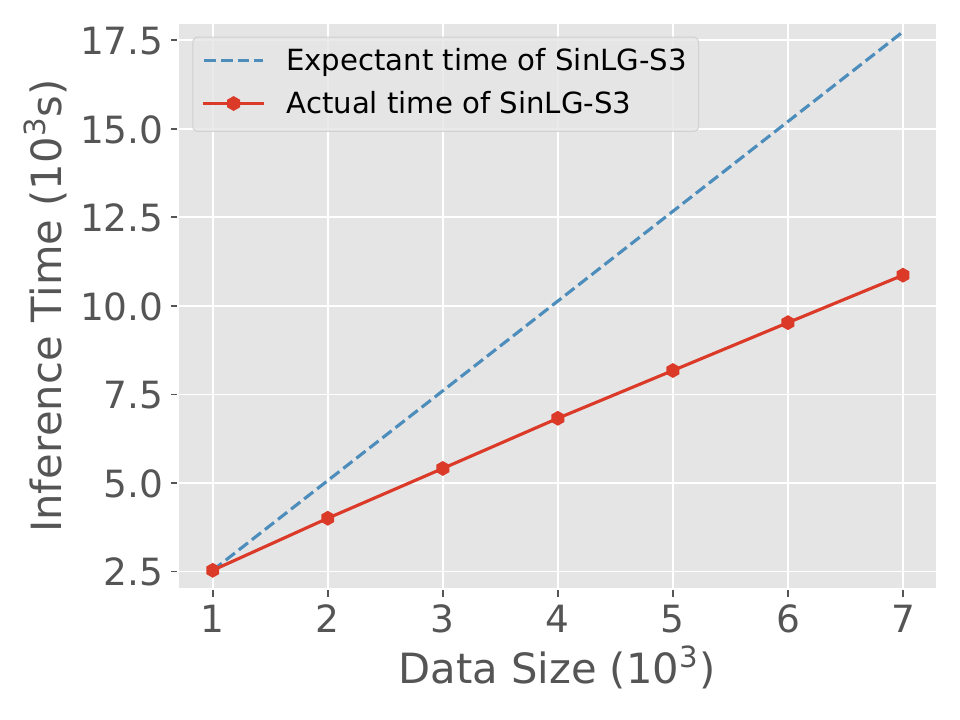}}
	\hspace{1ex}
	\subfigure[SinLG]{
		\label{fig:eva3:sub2}
		\includegraphics[width=0.4\textwidth]{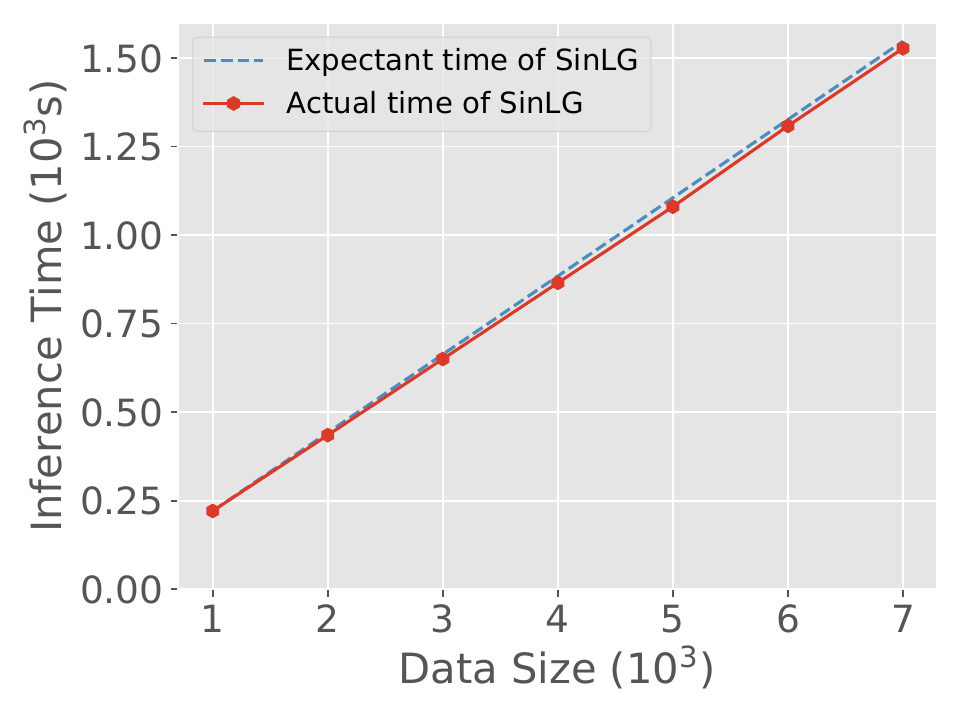}}
	\caption{Efficiency analysis of models. Note that the inference time corresponds to $1\sim7 \times 10^3$ instances. For each instance with $20$ response candidates, the average, worst, and best time costs of SinLG-S3 are $1.8315s$, $2.534s$, and $1.5523s$ while those of SinLG are $0.2171s$, $0.221s$ and $0.216s$, respectively.}
	\label{fig:eva3} 
\end{figure*}

To validate the performance gain from each component of SinLG, we conduct the ablation study which means cutting off some components from SinLG each time to form a simple-version model and check its performance via experiments. In other words, increasing our devised components step by step from the base model is also an alternative way to verify their effectiveness. We implement several simple variants of SinLG via the above ways, comparing with the PLM base model (Roberta) and SinLG (Roberta-based):

\begin{itemize}
    \item{\textbf{
    SinLG-S0:} A straightforward baseline, where we perform entity linking and then append the extracted commonsense knowledge to the input utterance directly. In this way, the GNN part is not in use.} 
    
	\item{\textbf{
 SinLG-S1:}} In this variant, we obtain the concatenation of the mean pooling results of concept embeddings and the context representation from Roberta, and feed it into an MLP layer to make the prediction. The concept embeddings are pre-computed via Roberta without fine-tuning.
	
	\item{\textbf{SinLG-S2:}} Although the implementation of SinLG-S1 seems simple and valid, it needs to query related knowledge from KG even online and this process is complicated and time-consuming, which will lead to high latency and bad experience for users. Referring to the auxiliary function of the GNN in our model, we design the second variant to fulfill the query-online-free target. Based on SinLG-S1, we add the similarity loss between the mean pooling results of concept embeddings and the context representation and conduct inference with Roberta only.
	
	\item{\textbf{SinLG-S3:}} Compared with SinLG, this variant omits the similarity loss part so that we can testify if the GNN part is useful in the whole architecture. The final matching scores are calculated according to the concatenation of representations from both the GNN and Roberta through an MLP layer.
\end{itemize}

In \textbf{Table \ref{tab:as1}} and \textbf{Table \ref{tab:as2}}, we summarize the components that each model contains besides giving the test results. Especially, SL represents similarity loss while QO-free means query-online- free, that is, there is no need to extract related concepts from the knowledge graph when the model conducts inference online or on validation and test datasets. The time cost with/without QO-free (i.e., SinLG-S3 and SinLG) is given in \textbf{Fig.\ref{fig:eva3}}.
\begin{itemize}
	\item According to the results of SinLG-S1, we can observe that the commonsense knowledge extracted from ConceptNet is useful for improving the performance of the model because it gains obvious promotion on both datasets. This is consistent with our assumption at the beginning. Compared with the other variants' performance gain, SinLG-S1's is the largest one, which indicates that the most effective part of our model is the incorporation of commonsense knowledge from KG. The simple operation, concatenation of the knowledge embeddings' mean pooling, and the representation of the PLM can improve by about 2 points.
	\item Compared with SinLG-S1, the results of SinLG-S2 are regressed on both PERSONA-CHAT original and revised, which illustrates that the combination of similarity loss and the SinLG-S1 is not that suitable. The straightforward operation, i.e., mean pooling could not be able to strengthen the knowledge signal from external KG sufficiently.
	\item From the result comparisons of SinLG-S3 and SinLG, it is obvious that the model's performance even improved with the similarity loss between representations from the GNN and PLM. The augmented representations from GNN can arouse the PLM's related memories more effectively so that it can attain better performance. The possible reason is that GNN can enhance the representation of relevant knowledge via message passing.
	\item As \textbf{Fig.\ref{fig:eva3}} shows, the inference time is correspond to $1\sim7 \times 10^3$ instances. For each instance with $20$ response candidates, the average, worst, and best time costs of SinLG-S3 are $1.8315s$, $2.534s$, and $1.5523s$ while those of SinLG are $0.2171s$, $0.221s$ and $0.216s$, respectively. Both SinLG-S3 and SinLG exhibit linearly-growth time costs with the increase of data, but with the QO process, SinLG-S3 needs to consume $7\sim11$ times SinLG's cost. This enlightens us that the similarity loss component is very desirable especially when the online delay is considered. 
\end{itemize}
Through the ablation study, we can draw a conclusion that the most effective part of our model is the fusion of extra common sense from the knowledge graph while GNN is also a strong assistant for the query-online-free target.

\begin{figure*}[tb!]
	\centering
	\setlength{\abovecaptionskip}{0.15cm}
	\subfigure[$R_{20}@1$]{
		\label{fig:eva4:sub1}
		\includegraphics[width=0.4\textwidth]{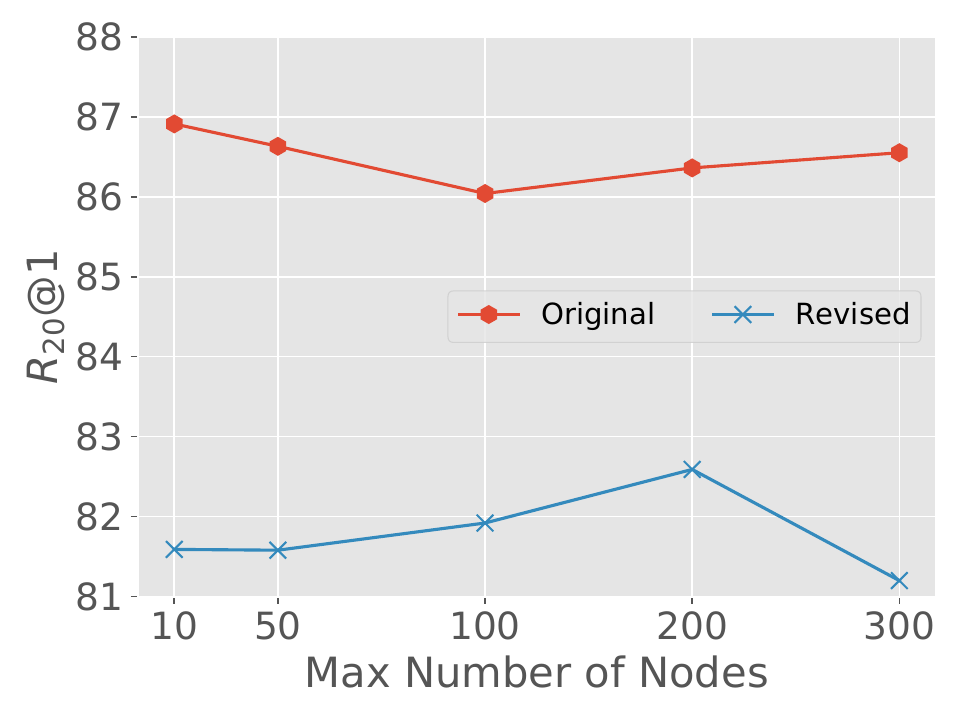}}
	\hspace{1ex}
	\subfigure[$R_{20}@2$]{
		\label{fig:eva4:sub2}
		\includegraphics[width=0.4\textwidth]{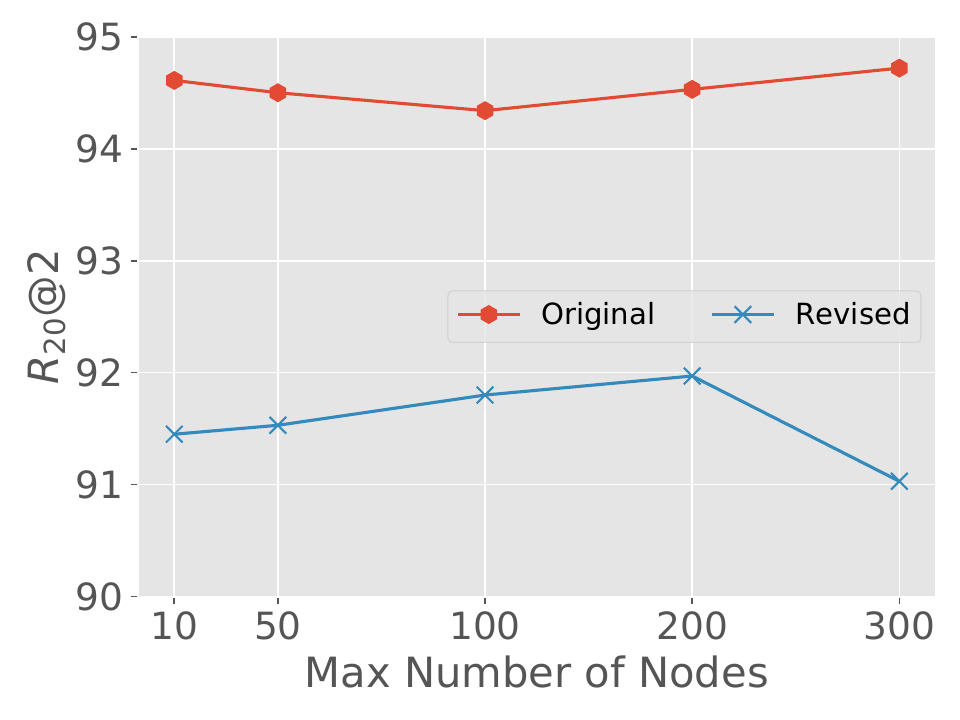}}
	\hspace{1ex}
	\subfigure[$R_{20}@5$]{
		\label{fig:eva4:sub3}
		\includegraphics[width=0.4\textwidth]{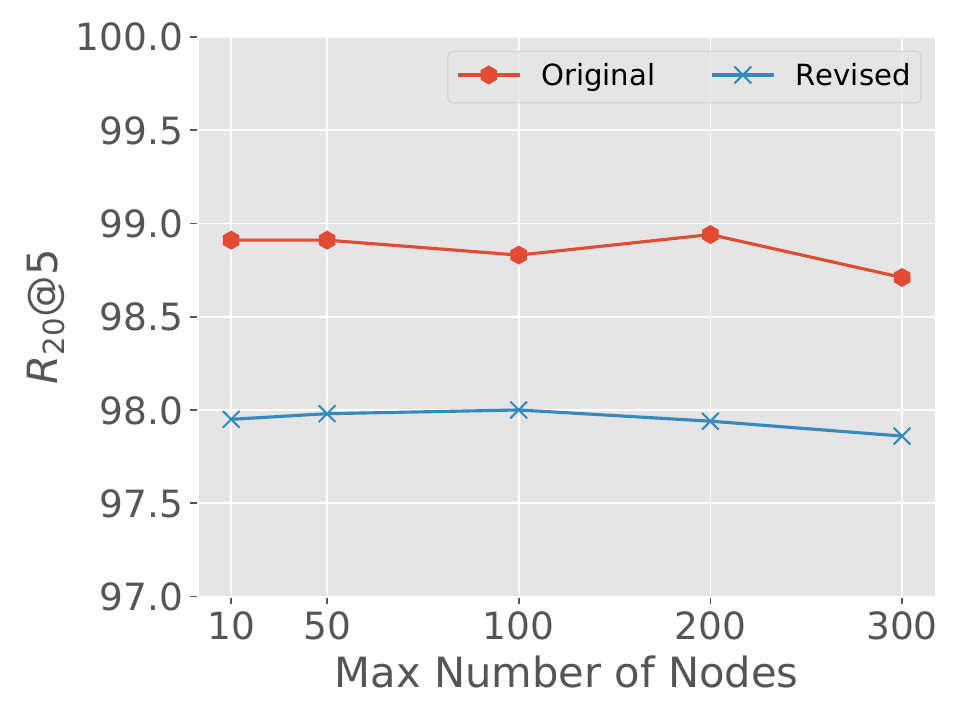}}
	\hspace{1ex}
	\subfigure[MRR]{
		\label{fig:eva4:sub4}
		\includegraphics[width=0.4\textwidth]{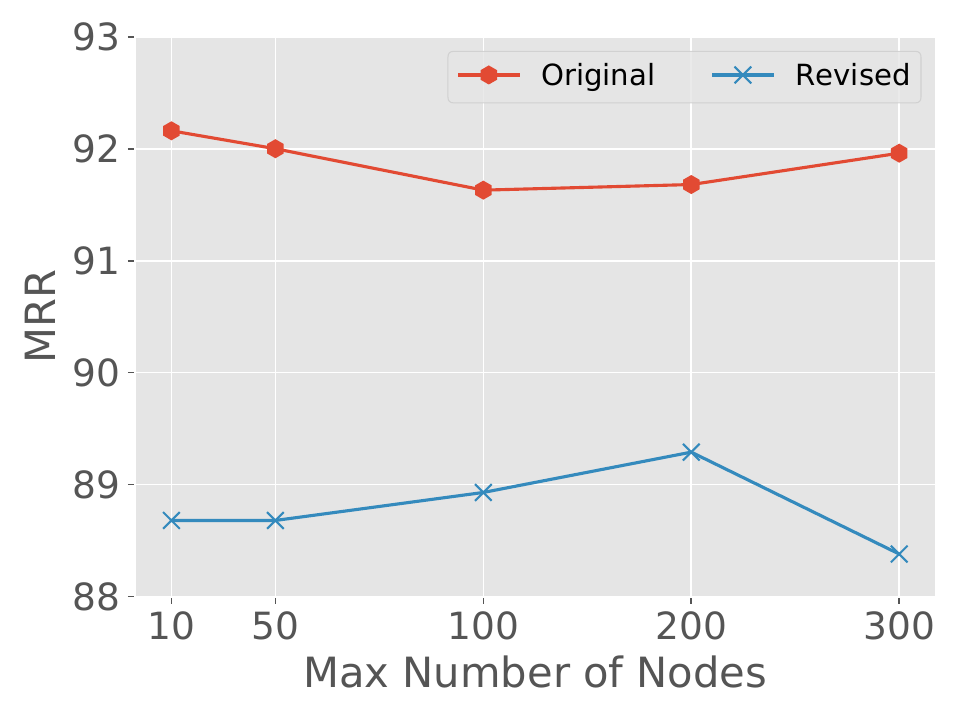}}
	\caption{SinLG performance under different maximum node numbers.}
	\label{fig:eva4} 
\end{figure*}

\begin{figure*}[tb!]
	\centering
	\setlength{\abovecaptionskip}{0.15cm}
	\subfigure[$R_{20}@1$]{
		\label{fig:eva5:sub1}
		\includegraphics[width=0.4\textwidth]{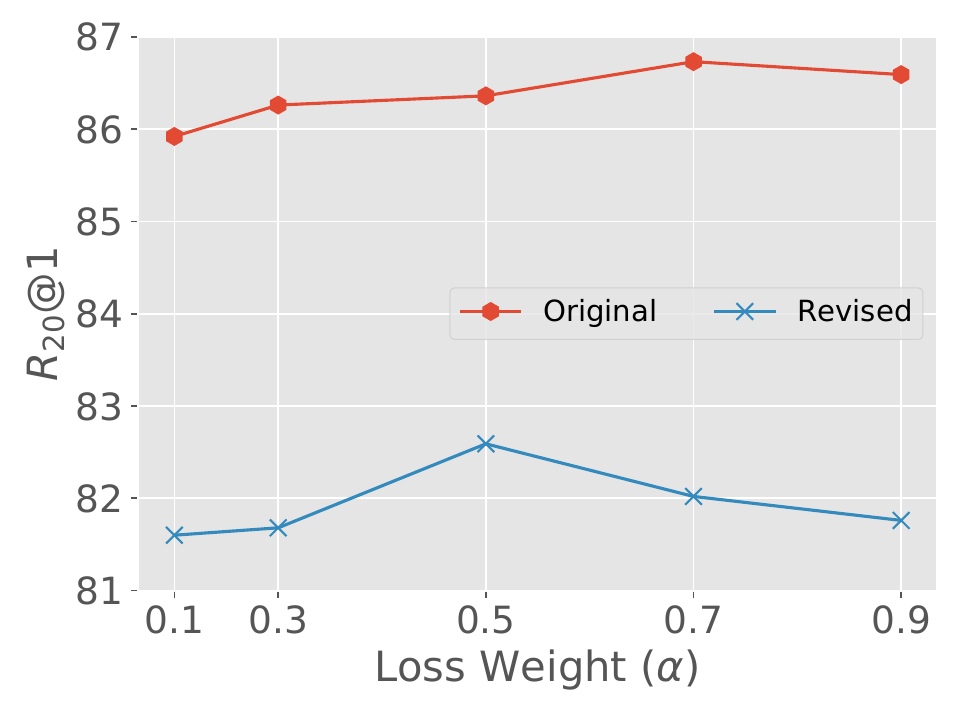}}
	\hspace{1ex}
	\subfigure[$R_{20}@2$]{
		\label{fig:eva5:sub2}
		\includegraphics[width=0.4\textwidth]{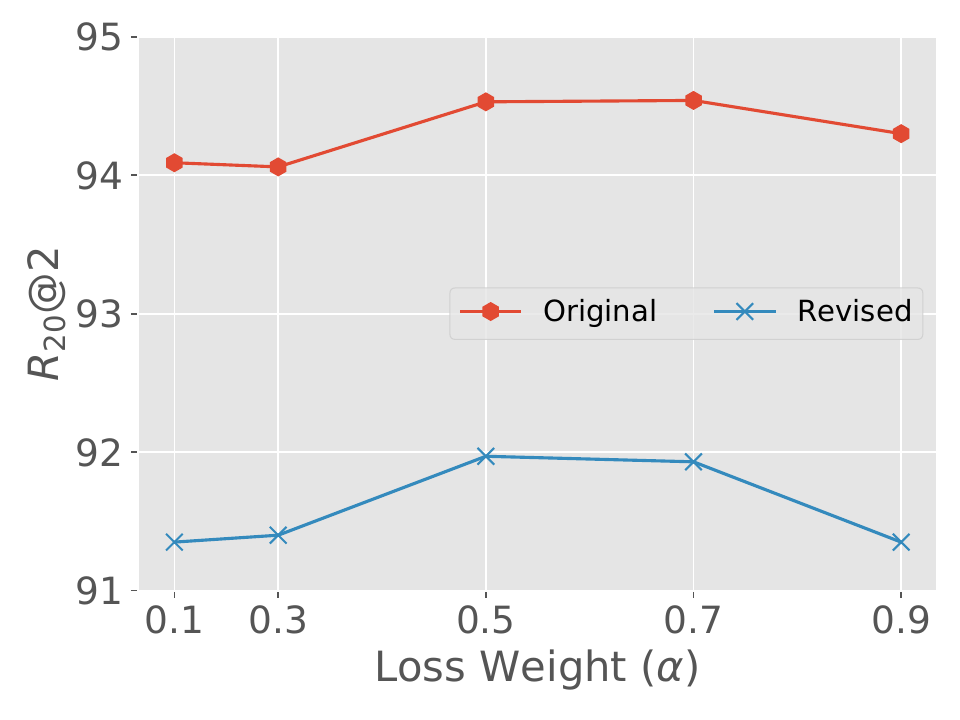}}
	\hspace{1ex}
	\subfigure[$R_{20}@5$]{
		\label{fig:eva5:sub3}
		\includegraphics[width=0.4\textwidth]{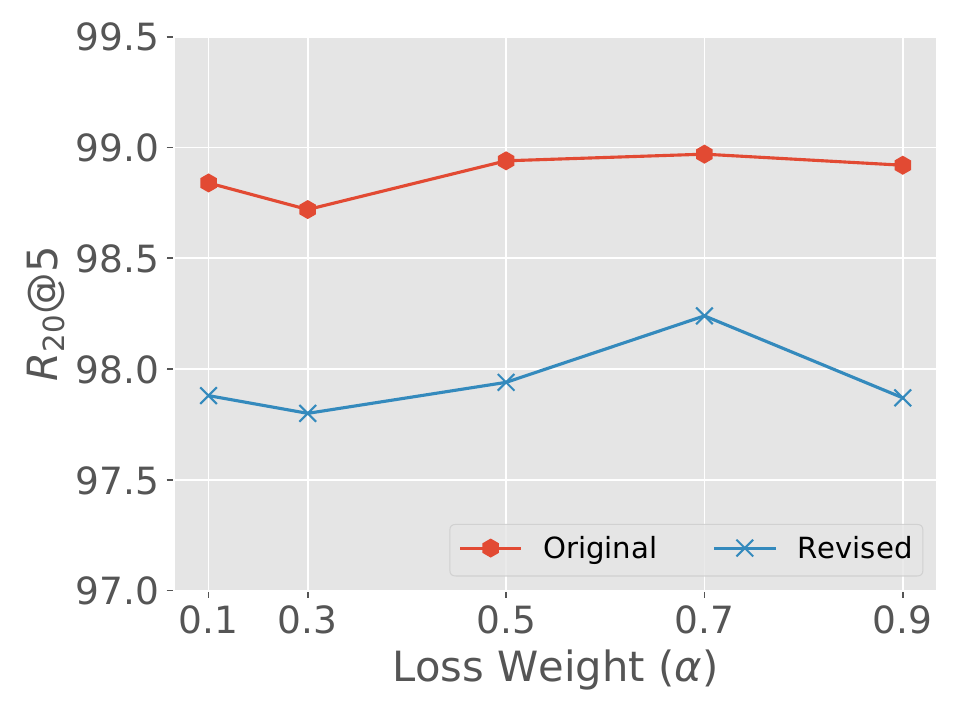}}
	\hspace{1ex}
	\subfigure[MRR]{
		\label{fig:eva5:sub4}
		\includegraphics[width=0.4\textwidth]{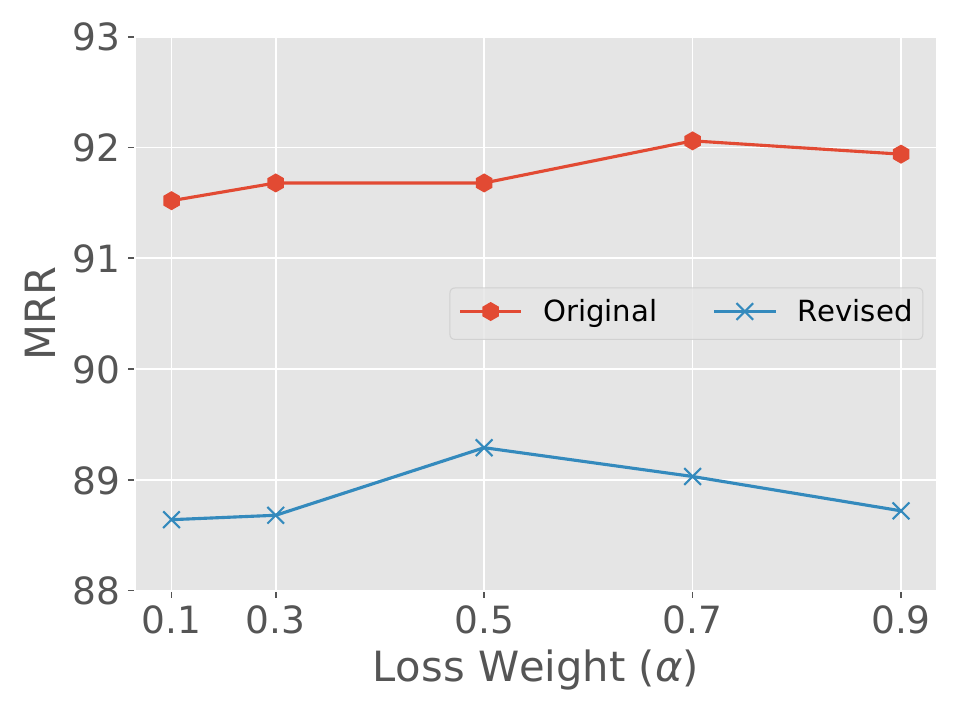}}
	\caption{SinLG performance under different loss weights.}
	\label{fig:eva5} 
\end{figure*}

\subsubsection{\textbf{The Analysis of Hyperparameter Sensitivity (RQ5).}}\label{sec:analysis}
In this section, we would show the performance changing of our model when the hyperparameters choose different values. This section discusses two worth-discussing hyperparameters which are the maximum number of concept nodes to construct the subgraph $\mathcal{G'}^s_{i,k}$, i.e., $\mathcal{V'}_{i,k}$ (10, 50, 100, 200, 300) in \textbf{Eq.(\ref{eq:transb})} and the loss weight $\alpha$ (0.1, 0.3, 0.5, 0.7, 0.9) in \textbf{Eq.(\ref{eq:final-loss})}. 

\paragraph{\textbf{Maximum node number}}
\begin{itemize}
	\item From \textbf{Fig.\ref{fig:eva4}}, we can observe that the best maximum node numbers for the two variant datasets are different. The model achieves relatively better performance on PERSONA-CHAT original when the maximum node number is set as 10 and the model performance decreases from 10 to 100 and then increases a bit between 100 and 300. According to the analysis, we consider there are two possible reasons that lead to this phenomenon. First, PERSONA-CHAT original is a simpler dataset whose persona information is so easy to understand for PLMs that not much extra commonsense knowledge is needed. Second, the concepts extracted from KG between 10 to 100 contain some impactive noises, which leads to the performance drop. 
	\item From the blue line with crosses in \textbf{Fig.\ref{fig:eva4}}, it is obvious that our model performs best on PERSONA-CHAT revised when the maximum node number is set as 200. As the maximum node number determines the message passing distance in the GNN part and the amount of extra commonsense knowledge from KG, it is reasonable that this depends on the attributes of specific datasets.
\end{itemize}

\paragraph{\textbf{Loss weight $\alpha$}}
\begin{itemize}
	\item According to the curve of the red line with dots in \textbf{Fig.\ref{fig:eva5}}, our model obtains the best performance on PERSONA-CHAT original when the loss weight is 0.7, which demonstrates that the binary cross entropy plays a more important role than the cosine similarity in the loss function. This is consistent with the assumption that we mentioned above that PERSONA-CHAT original is easier and not much extra commonsense knowledge is needed.
	\item According to the curve of the blue line with crosses in \textbf{Fig.\ref{fig:eva5}}, our model achieves the best performance when the loss weight is set as 0.5, which indicates that our proposed cosine similarity has an equal influence on the final results on PERSONA-CHAT revised. This testifies the effectiveness of extra commonsense knowledge for difficult understanding tasks.
\end{itemize}

In summary, our model is sensitive to these two hyperparameters which depend on the difficult levels of target datasets. Both the performance rangeability of our model and the optimal values of hyperparameters would be affected by the difficult levels of target datasets.

\section{State-of-the-art}\label{sec:related}
In this section, we introduce existing studies from two categories: the development of models on MRS, and some recent knowledge reasoning methods that provide us with insights and inspirations in this work.
\vspace{-2ex}
\subsection{Methods on Multi-turn Response Selection}
In order to solve the multi-turn response selection problem, existing studies propose various context-response matching models which include traditional methods and \ac{PLM}-based ones~\cite{tao2021survey}.

Traditional methods first focus on learning better embeddings for both the context and response. The concatenation of them is fed into a matching layer to make the final prediction~\cite{wang2017deep,wu2018response,yan2018response,xu2021topic}.
Specifically,~\citet{wang2017deep} utilize hand-craft features, i.e., sentence length, the number of common words for sentence-level representation while~\citet{wu2018response} employ topic features to provide clues for the selection of the best response. 
With the prosperity of deep neural networks, researchers propose employing \ac{CNN}, \ac{RNN}, and self-attention mechanism, combining with \ac{MLP}, pooling layers to learn effective representations. For instance,~\citet{yan2018response}  extract features from three different levels, that is, word, phrase, and sentence and propose an attention-based \ac{CNN} model to match the dialogue context with response candidates. 
Different from the previous methods that use topic-agnostic n-gram utterances as processing elements,~\citet{xu2021topic} propose capturing topic shift at a discourse level and then effectively tracking the topic flow in the multi-turn dialogue. 

Then the interactions between context and response draw intensive attention. Diverse new frameworks~\cite{wu2017sequential, wu2018knowledge, tay2018co, zhou2018multi, tao2019one, yuan2019multi} are presented to mine the interaction between them so that the matching accuracy can be improved. 
The study~\cite{tay2018co} designs the CARAN, stacked 
 recurrent encoders, which consist of a bidirectional alignment and a multi-level attention refinement mechanism.
Inspired by the rationale of Transformer,~\citet{zhou2018multi} propose a deep network based entirely on attention. Specifically, they create embeddings of context utterances at different granularities and then distill the highly-relative segment pairs with attention scores.
Previous studies usually conduct interactions between context and response only on time, but~\citet{tao2019one} raise this is conducted in a shallow way. Hence, they propose an interaction-over-interaction network to further the context-response interaction. 
Typically,~\citet{yuan2019multi} analyze the negative effects of involving exceeded context sentences and propose MSN to filter out the most relative sentences. 
As expected, the above methods keep trying to go deep into interactions fully and even among each word within the context while this can be accomplished well via PLMs.

Since the release of the first \ac{PLM}~\cite{devlin2019bert}, various impressive results on downstream \ac{NLP} tasks are achieved due to its strong capability in language representation and understanding. PLM-based models~\cite{henderson2019training, whang2020effective, wu2020tod, xu2021learning, song2021bob} for MRS also emerge and lead to outstanding performance. Typically, the study~\cite{henderson2019training} learns embedding for both context and response severally via \ac{BERT} and computes the dot product of them as the final matching score.
Although BERT is conveniently adapted to various NLP tasks, ~\citet{whang2020effective} believe it still has limitations for tasks on a certain domain so they first post-train BERT on their task-specific corpora with two objectives, i.e., Masked Language Model (MLM) and Next Sentence Prediction (NSP), and then fine-tune BERT on the response selection task to achieve a better performance than the previous state-of-the-art.
Considering the generalization of PLMs and the particularity of specific tasks,~\citet{wu2020tod} propose TOD-BERT that incorporates speaker and system tokens into the masked language modeling during pre-training to model the dialogue attributes better.
Further,~\citet{xu2021learning} design several self-supervised tasks to pre-train the PLM-based model in a multi-task manner for MRS where remarkable results have been obtained.

Meanwhile, external knowledge in paragraphs (such as profiles of speakers, and entity infobox from Wikipedia) is employed to assist in solving MSR problem~\cite{zhao2019document, gu2019dually, gu2020filtering, liu2020k, gu2021partner, zhang2021adapting}. To overcome the challenge of grounding dialogue contexts with background texts and distinguishing significant information, DGMN~\cite{zhao2019document} encodes three contents (i.e., sentences from the document, utterances in the dialogue context, and the target response candidate) simultaneously via self-attention, and contains attention mechanism for document-aware contexts and context-aware documents so that it can learn a rich representation for both dialogue contexts and candidate responses.
Different from existing persona fusion approaches that enhance the context representation by calculating the similarity between it and a predefined persona,~\citet{gu2019dually} propose DIM which ranks response candidates via interactive matching between responses and contexts, as well as responses and personas respectively.
In order to make full use of the background knowledge, and simultaneously match response candidates with both the context and its knowledge,~\citet{gu2020filtering} propose FIRE. They build two filters for the knowledge and the context, which generate representations for both knowledge and context under the other's ground. 
\citet{liu2020k} incorporate knowledge triplets into the utterances and feed them into BERT while they use soft position and visible matrix to optimize the knowledge noise issues. 
\citet{gu2021partner} explore the effect of utilizing persona information based on the PERSONA-CHAT dataset and design four persona fusion strategies which are implemented into three existing models to testify their effectiveness.
However, the above studies haven't considered the effectiveness of common sense from extra knowledge graphs so we propose a model to incorporate this information and assist the training of PLMs.

\subsection{Knowledge Reasoning Methods}
As a prevalent and effective method~\cite{chen2019towards, lin2019kagnet, du2021cogkr, zhang2022subgraph, yu2022a} in natural language processing and knowledge discovery domain, knowledge reasoning provides us with many insights for our work. 
Especially,~\citet{lin2019kagnet} present an inference framework for commonsense question answering, which first projects the question-answer text from the semantic way to the knowledge-based symbolic way, forming a subgraph of an external knowledge graph. Then, it embeds schema graphs with a network based on GCN, LSTM, and a hierarchical attention mechanism. At last, it scores each answer conditioned on graph representations. The intermediate attention scores lead to transparent, interpretable, and trustworthy inferences. 
\citet{chen2019towards} put forward KBRD, which capacitates interactions between recommender and dialogue systems. In this framework, informative entities are connected to an extra knowledge graph and input into the recommender besides items and they are transmitted on the KG through a relational graph convolutional network to enrich the user interest's representation. Moreover, the knowledge-enhanced representation is put back to the dialogue system as a form of vocabulary bias, capacitating it to select responses that match the user's interest.
Few previous studies pay attention to the relation types in the knowledge graph when extracting new facts from them, especially only one or few instances are given,~\citet{du2021cogkr} propose CogKR, which summarizes the underlying relations of the provided instances first and then builds a cognitive graph through coordinating retrieval and reasoning iteratively.
Existing subgraph-retrieval methods from KG tend to increase reasoning bias, so~\citet{zhang2022subgraph} design a trainable subgraph retriever separated with the following reasoning process, which can broaden paths to induce the subgraph and stop automatic expansion. Above this, any subgraph-oriented reasoners can be utilized to delicately deduce more reasonable answers from the subgraph.
\citet{yu2022a} design a knowledge-grounded dialogue system, called XDAI \footnote{https://github.com/THUDM/XDAI} which possesses the prompt-aware PLM exploitation without fine-tuning and is equipped with open-domain knowledge and domain-specific mechanism. It is convenient for developers to quickly create an open-domain or domain-oriented dialogue system without heavy fine-tuning of PLMs.

Inspiring by the dual process theory from the cognitive process
of humans~\cite{evans1984heuristic, evans2003two, evans2008dual, sloman1996empirical}, frameworks~\cite{ding2019cognitive, feng2020scalable, lv2020graph, yasunaga2021qa, liu2021graph, zhang2021greaselm, zhou2022eventbert} combining \ac{PLM} with \ac{GNN} are presented to enhance both the reasoning ability and interpretability of intelligent systems.
Specifically,~\citet{ding2019cognitive} propose CogQA for multi-hop question answering with the help of web-scale documents, which iteratively constructs a cognitive graph through the implicit extraction module (\ac{BERT}) and conducts explicit reasoning via \ac{GNN}. 
\citet{yasunaga2021qa} present QA-GNN, another PLM+KG model for question answering, which first encodes the QA context via a PLM and then retrieves a KG subgraph that includes all multi-hop neighbor entities of QA. As some entities are always more relevant to the QA, a PLM is directly used to score them. In addition, the QA context is regarded as a super node of its KG subgraph, and its representation from PLM is the initial feature. Finally, the representation of the super node via graph attention network and PLMs, as well as the subgraph's representation are concatenated together to obtain the inference through an MLP layer.
\citet{liu2021graph} propose GRN which pre-trains an ALBERT with two self-supervised tasks and then fine-tunes it with a graph reasoning module together.
Beyond combining PLM and GNN simply,~\citet{zhang2021greaselm} present GREASELM which fuses and exchanges information from all the PLM and KG in multiple layers. An interactive mechanism to bi-directionally exchange information is designed between each layer of the PLM and GNN. Therefore, the inputs of two modalities can interact with each other directly.

The studies introduced in this section not only provide us with many inspirations for our current work but also enlighten us about prospects in relevant domains.

\section{CONCLUSION}\label{sec:conclusion}
In recent years, dialogue systems are attracting more and more attention from the public. MRS is a general research problem during developing a dialogue system. Combining the practice of the dialogue system, we find three main challenges of MRS. How to comprehensively understand the relationships between different utterances in the context, response candidates, and background information under the multi-turn scenario; how to select a rational response consistent with common sense; how to achieve a performance gain even under the low-resource scenario. In this work, we propose the framework SinLG where Pre-trained Language Models (PLMs) aim to understand the language correlation among different utterances in context, response candidates, and background information while the Graph Neural Network (GNN) is responsible for reasoning useful commonsense information from the extra knowledge base and assists PLMs in fine-tuning. We testify the enhancing effects of common sense from the knowledge graph and the GNN to PLMs on the public dataset, which demonstrates that our framework can not only improve PLMs' performance on tasks with different levels of understanding difficulty but also achieve more performance gain under the low resource scenario. 

However, one sufficiency of our model is the operation of the dialogue context is too coarse. If the length of the conversation extends to more than the maximum length of the model, some information may be lost. In the future, we can design a more reasonable and practical component to catch the whole dialogue's information as much as possible. For example, we can learn the relationships of different sentences, get rid of the useless parts, and simplify the conversation before inputting the model.

\begin{acks}
	The authors would like to thank Ming Ding and Yangli-ao Geng for their empirical suggestions on this work. The authors would also like to thank the anonymous reviewers for their insightful comments provided. 
	
\end{acks}

\bibliographystyle{plainnat}
\bibliography{MRS}

\end{document}